\documentclass[11pt,letterpaper]{article}
\usepackage[numbers]{natbib}
\usepackage[margin=1in]{geometry}
\usepackage{graphicx}
\usepackage{amsmath}
\usepackage{float}
\usepackage{placeins}
\usepackage{subcaption}
\usepackage{authblk}
\usepackage[hyphens]{url}  
\usepackage[hidelinks=false,breaklinks=true]{hyperref}
\usepackage{booktabs}

\title{Into the crossfire: evaluating the use of a language model to crowdsource gun violence reports}

\author[1]{Adriano Belisario}
\author[1]{Scott A.\ Hale\thanks{These authors contributed equally to this work and share senior authorship.}}
\author[1]{Luc Rocher\unskip\textsuperscript{*}}

\affil[1]{Oxford Internet Institute, University of Oxford, Oxford, UK}

\date{\today}

\begin{document}

\maketitle
\renewcommand{\thefootnote}{}
\footnotetext{
© Adriano Belisario, Luc Rocher, Scott Hale 2025. 
This is the author's version of the work. It is posted here for your personal use. 
Not for redistribution. The definitive Version of Record was published in 
\textit{Proceedings of the ACM on Human-Computer Interaction}, 
Vol.\ 9, No.\ 7, CSCW235, November 2025. 
\href{https://doi.org/10.1145/3757416}{https://doi.org/10.1145/3757416}.
}
\renewcommand{\thefootnote}{\arabic{footnote}}
\begin{abstract}
Gun violence is a pressing human rights issue that affects nearly every dimension of the social fabric, from healthcare and education to psychology and the economy. Reliable data on firearm events is paramount to developing more effective public policy and emergency responses. However, the lack of comprehensive databases and the risks of in-person surveys prevent human rights organizations from collecting needed data in most countries. Here, we partner with a Brazilian human rights organization to conduct a systematic evaluation of language models to assist with monitoring real-world firearm events from social media data. We propose a fine-tuned BERT-based model trained on Twitter (now X) texts to distinguish gun violence reports from ordinary Portuguese texts. We then incorporate our model into a web application and test it in a live intervention. We study and interview Brazilian analysts who continuously check social media texts to identify new gun violence events. Qualitative assessments show that our solution helped all analysts use their time more efficiently and expanded their search capacities. Quantitative assessments show that the use of our model was associated with analysts having further interactions with online users reporting gun violence. Our findings suggest that human-centered interventions using language models can help support the work of human rights organizations.
\end{abstract}


\section{Introduction}

\label{chap:intro}

Numbers can hardly convey the tragic losses caused by gun violence. Yet they provide valuable insights to understand and address pressing violations of human rights. According to a comprehensive international study from 2022~\citep{ouGlobalBurdenTrends2022}, cases of physical violence by firearm have increased over the past three decades. These cases are heterogeneously distributed across countries, with nations such as Brazil and the USA accounting for a considerable share of the global burden of firearm violence~\citep{globalBurden2018,ouGlobalBurdenTrends2022}. The scale of this issue has led to firearm violence being described as an epidemic~\citep{szwarcwaldMortalidadePorArmas1998,kalesanHiddenEpidemicFirearm2017,cavalcantiDinamicasViolenciaUrbana2017,fontanarosaUnrelentingEpidemicFirearm2022} and a public health crisis~\citep{ezeImpactGunViolence2023,silverExaminingHealthcareEconomic2023,generalFirearmViolenceAmerica2024} in these countries. Furthermore, beyond loss of life, this burden impacts nearly every dimension of the social fabric, such as healthcare~\citep{lozovatskyImpactFirearmViolence2014,silvaNoMeioFogo2021}, education~\citep{lemgruberTirosNoFuturo2022}, psychology~\citep{garbarinoMitigatingEffectsGun2002}, and the economy ~\citep{silverExaminingHealthcareEconomic2023,ouGlobalBurdenTrends2022}.

Reliable data on firearm events is paramount to developing more effective public policy and emergency responses, but documenting these cases is challenging. As comprehensive databases of human rights violations are rare~\citep{priceLimitsObservationUnderstanding2015}, humanitarian organizations are trialling social media data to supersede risky and costly studies on-site~\citep{mooneySocialMediaEvidence2021,koliebRespondingHumanRights2018}. They typically use keyword-based searches to crowdsource online evidence, which often results in large datasets with a high proportion of unrelated texts~\citep{myersHowConductDiscovery2020,koenigOpenSourceInvestigations2020}. Reviewing and filtering these datasets to find firearm violence reports require substantial manual work and can be akin to searching for a needle in a haystack.

To handle this problem, Natural Language Processing (NLP) methods have emerged as a promising solution to automate text classification, and researchers developed models to automate tasks stemming from social media data in quantitative human rights studies. For instance, classification models have been successfully developed to assist human rights investigations based on social media data in English~\citep{pilankarDetectingViolationHuman2022}, Spanish~\citep{arellanoOverviewDAVINCISIberLEF2022}, in the Arab world using Twitter data~\citep{alhelbawyNLPPoweredHumanRights2020}, and on the Russian-Ukraine war using Telegram messages~\citep{nemkovaDetectingHumanRights2023}. These studies suggest that NLP methods can achieve promising results for quantitative human rights classification tasks when evaluated on benchmark datasets. However, little is known about the effectiveness of NLP methods on live and dynamic social media data in real-world monitoring applications.

Here, we publish the first systematic real-world evaluation of a language model to assist human rights analysts in crowdsourcing gun violence reports from social media. We partnered with Fogo Cruzado Institute\footnote{Instituto Fogo Cruzado stands for ``Crossfire Institute'': \href{https://fogocruzado.org.br}{https://fogocruzado.org.br}.}, a human rights organization that monitors gun violence in Brazil, and produces real-time situational awareness alerts for local citizens. 

Fogo Cruzado continuously monitors online data, including Twitter posts, to identify gun violence events in four Brazilian regions. Analysts use Twitter to work in two stages: first, they monitor specific keywords and profiles. Second, once they find a potential gun violence report, they interact publicly with the Twitter user to collect further information and fact-check the event. We aim to automate and augment the first stage with NLP methods.

Our research sought to answer the following research questions:

\begin{itemize}
\item RQ1 - Can Transformer-based language models accurately identify gun violence reports in Brazilian Portuguese social media texts?

\item RQ2 - What are the advantages and challenges of adopting a language model for real-time monitoring compared to manually reviewing social media texts?
\end{itemize}

We first develop a Bidirectional Encoder Representations from Transformers (BERT)~\citep{devlinBERTPretrainingDeep2018} model to classify whether social media texts contain gun violence reports. We build upon BERTimbau~\citep{souzaBERTimbauPretrainedBERT2020}, an open-source BERT model pre-trained on Brazilian Portuguese text. To classify gun violence events, we fine-tune this model using Twitter texts with semi-supervised learning techniques. This leads to a model that can accurately classify whether a text message contains a gun violence report (positive cases) or not (negative) with a recall rate for positive cases of 87\% on our human-reviewed evaluation set (random baseline at 19\%).

We then develop a web application that uses BERTimbau to help human rights analysts identify gun violence reports from Twitter data in real time. Using an intervention design, we introduced Fogo Cruzado's analysts in Rio de Janeiro to our web application, and they used it between 28 May 28 and 2 July 2023.  Qualitative assessments show that our solution helped all analysts use their time more efficiently and expanded their search capacities. Quantitative assessments show that the use of our model was associated with analysts having more interactions with online users reporting gun violence. Taken together, our findings suggest that language models can significantly augment the ability of human rights analysts to monitor reports from social media.

\section{Related work}
\label{chap:related}
Data collection has long been part of human rights research, yet adopting quantitative methods for analysis is a recent development. Local and international organizations initially collected data from paper-based questionnaires, written testimonies, and surveys to register human rights abuses ~\citep{amnesty_international_amnesty_2011, mcclintock_standard_2010}. Later, they adopted digital databases, but data on human rights violations still lacked basic standardization procedures required for robust statistical analysis~\citep{ball_using_2019}. Addressing this problem, organizations and academic scholars began developing new approaches for data collection and analysis in the early 1990s~\citep{jabine_human_1992,ball_using_2019,murdie_quantitative_2021}. Among the drivers of the trend towards quantitative methods in human rights research, \citet{murdie_quantitative_2021} highlighted advancements in technology and data availability, enhanced methodologies and on-the-ground collaborations. 

The greater data availability for human rights research has partly occurred due to social media platforms. Compared to existing records, such as survivor testimonies~\citep{millerDiggingHumanRights2013,gokhaleDeployingCotrainingAlgorithm2017} or newspaper articles~\citep{pavlickGunViolenceDatabase2016,bauerNLPHumanRights2022,ranNewTaskDataset2023}, social media platforms have emerged as a ``live source'' for data collection in human rights studies, offering a rapid way to gather real-time data from a broad user base. Thus, crowdsourcing user-generated content on social media holds great potential for public interest technologies, which aim to benefit communities and promote citizen well-being, rather than serving solely private interests~\citep{abbasSocioTechnicalDesignPublic2021}. In this vein, researchers have used social media data to understand ongoing conflicts~\citep{koliebRespondingHumanRights2018} and tackle important societal issues, including misinformation~\citep{shabaniHybridMachineCrowdApproach2018} and disaster management~\citep{nielsenSocialMediaCrowdsourcing2024}. 

Over the last decade, many models and datasets have been developed that leverage textual data to investigate rights violations. Such datasets include the Gun Violence Database from US daily news~\citep{pavlickGunViolenceDatabase2016} or the Arabic Violence Twitter Corpus~\citep{alhelbawyCorpusViolenceActs2016}. Similarly, human rights studies, particularly since the 2010s, have adopted NLP methods to analyze large volumes of text data, applying them for tasks such as named entity recognition~\citep{gokhaleDeployingCotrainingAlgorithm2017} and text classifiers~\citep{hu_conflibert_2022,alhelbawyNLPPoweredHumanRights2020,gokhaleDeployingCotrainingAlgorithm2017}. 

Advances in Machine Learning, including the introduction of the Transformer architecture~\citep{vaswaniAttentionAllYou2017}, have drastically changed the landscape of NLP applications in many fields. The Bidirectional Encoder Representations from Transformers (BERT) variants stood out as a promising solution for human rights violation event detection~\citep{nemkovaDetectingHumanRights2023,hu_conflibert_2022,taGANBERTAdversarialLearning2022,alhelbawyNLPPoweredHumanRights2020}. \citet{pilankarDetectingViolationHuman2022} used a BERT model to classify tweets with ``factual posts'' (as opposed to opinionated messages) about ``violation of human rights in any part of the world.'' 
Another example is ConfliBERT, a pre-trained BERT model for the political conflict and violence domain~\citep{hu_conflibert_2022}. This language model was trained using specialized vocabulary, and the authors claimed that ``ConfliBERT outperforms BERT when analyzing political violence and conflict''.

Human rights research on large-scale data is predominantly performed in English~\citep{hu_conflibert_2022, pavlickGunViolenceDatabase2016}. Applied research in low-resourced languages within NLP remains an emerging area of interest. In Spanish, ~\citet{taGANBERTAdversarialLearning2022} presents the findings of multiple studies using social media to detect violent incidents. All studies used variations of pre-trained Transformer models, including a wide variety of BERT implementations. In Russian, multiple BERT-based models have been used to classify Telegram messages and detect human rights violations during the Russian-Ukraine war~\citep{nemkovaDetectingHumanRights2023}. To the best of our knowledge, there has been no previous work on human rights abuse detection using social media texts in Portuguese.

To date, there is limited evidence of NLP models being evaluated for human rights monitoring in real-world settings. One example of such applied research is presented by \citet{alhelbawyNLPPoweredHumanRights2020}. The authors built a corpus of tweets reporting violent acts in Arabic~\citep{alhelbawyCorpusViolenceActs2016} and compared different NLP architectures, including baseline models (Naive Bayes and Support Vector Machine) and two types of long short-term memory (LSTM) neural networks. The LSTM architectures achieve the highest performance, while the authors note that Transformer-based architectures could offer higher scores. \citet{alhelbawyNLPPoweredHumanRights2020} concisely comment on the impact of its real-world implementation, stating that ``collected reports contributed to a number of publications by human rights organizations.'' However, there remains an important gap in systematically assessing the effectiveness of machine learning models for real-time detection of human rights abuse.

Public interest technologies applied to human rights often involve creating user-friendly human-computer interfaces to interact with and explore the data. For example, \citet{alhelbawyNLPPoweredHumanRights2020} integrated the results of the NLP classifier into a web platform, and \citet{millerDiggingHumanRights2013} used graphs to represent the relationship between entities in documents related to human rights violations. Nevertheless, knowledge about the challenges and effectiveness of using these interfaces for real-world monitoring remains scarce. To address this uncertainty, we propose a web-based interface that allows analysts to visualize model classifications of crowdsourced data and leverage survey data from human rights analysts who actively use our interface, providing valuable insights into the design and development of effective human-computer interfaces in this field.

Despite the potential of recent NLP methods and social media data, human rights organizations often struggle to effectively harness these technologies. Extracting actionable information from social media data demands specialized skills and expertise that many nonprofits lack~\citep{farmerDataSocialGood2023}. Additionally, nonprofits, including human rights organizations, often perceive social media primarily as a one-way communication tool to broadcast their agendas, rather than as a means for engagement or data collection~\citep{namisangoWhatWeKnow2019}. This technological and methodological gap is especially detrimental in fragile contexts, where timely access to accurate information is most needed to respond to crises and protect vulnerable populations.

\section{Background and context}
\label{chap:background}
Rio de Janeiro is systematically affected by firearm violence~\citep{silvaNoMeioFogo2021,lemgruberTirosNoFuturo2022}, but there are no granular official records of these events publicly available. The lack of microdata about gun violence events is critical and hinders academic research and society's capacity to produce evidence that informs public policy. The Mortality Information System offers national microdata about deaths and has been used by public security studies on lethality ~\citep{silvaNoMeioFogo2021,buenos_anuario_2023}, but not all gun violence implies deaths. Records from police stations, which could provide more details on events of armed violence, are not publicly available. The local government offers only aggregated public security metrics per month and administrative units. 

Fogo Cruzado is a non-governmental organization (NGO) founded in 2016 out of the need for granular data about the firearm violence cases in Rio. Initially incubated by Amnesty International Brazil, it now has nearly twenty professionals distributed across four metropolitan areas. Fogo Cruzado plays a significant role in advocacy, academic research and public awareness about gun violence. Internationally, it is an important data source for ACLED, the Armed Conflict Location and Event Data Project~\citep{raleighIntroducingACLEDArmed2010}, which gathers disaggregated data on armed violence worldwide and has been extensively used in academic research ~\citep{bloem_covid-19_2021,piskorski_new_2020}. In Brazil, Fogo Cruzado co-authored several academic reports on Rio de Janeiro's public security, including an analysis of police raids with high rates of lethality~\citep{hirataChacinasPoliciaisNo2023}, robberies~\citep{hirataRoubosProtecaoPatrimonial2019} and an extensive mapping the territorial control of armed groups in Rio de Janeiro~\citep{hirataMapaHistoricoDos2022}.

We conducted five video interviews with Fogo Cruzado's staff between May and July 2023, and this section builds upon the information gathered from these meetings. Fogo Cruzado employs a dedicated team of analysts to monitor events in each of the four regions monitored. The core team for monitoring gun violence events in Rio de Janeiro has four analysts, a team leader, and an additional position that rotates among other Brazilian states as needed. They work in shifts to record information related to gun violence events, such as the number of dead or injured civilians and police forces and whether there was an ongoing police operation. The team monitors new cases seven days a week, except in the period between midnight and 6 am. The analyst working early in the morning is in charge of retrospectively catching up and recording the cases reported during this night period. They receive reports submitted by citizens using Fogo Cruzado's mobile application and actively search for reports on social media platforms. Analysts follow local WhatsApp groups, Facebook pages and groups, Twitter posts, and local press websites. After identifying a report, the team checks the event cross-verifying it with other online and on-the-ground sources.

Fogo Cruzado has systematically used Twitter to track and interact with gun violence reports since 2018. This social network is an ``essential platform for research'', according to Fogo Cruzado's Systematization Protocol.\footnote{Fogo Cruzado's internal document kindly shared for this research.} Fogo Cruzado uses a different Twitter profile to track and interact with users reporting gun violence events in each region. We requested the team in Rio de Janeiro a detailed account of the sources for identifying new cases. They recorded the primary source of 150 events over the period from 2 February 2023 to 15 March 2023. Twitter was used in 68\% of these events recorded and fact-checked by the team. WhatsApp was the second most important source, accounting for 35\% of cases. Fogo Cruzado's mobile application contributed 5\% of cases, while Facebook and personal contacts each played a minor role, accounting for less than 1\% individually. Percentages do not add up to 100\% since the same event can have multiple sources. Although limited, these results confirm the relevance of Twitter as a data source for identifying gun violence reports in real time. 

Some of the keywords monitored on Twitter (``\textit{(bala voando) OR tiro OR tiroteio OR baleado}”\footnote{Verbatim translation: ``\textit{(bullet flying) OR shot OR shooting OR [person] shot}”}) are strongly associated with shooting events, but others are common terms in Portuguese. While wordings such as ``bala voando” (``bullet flying”) are literal representations of firearm events, the word ``tiro” not only means ``shot” but is also the first person present tense for the verb ``tirar” (``to take”). Consequently, it is primarily used in various contexts unrelated to gun violence reports. Furthermore, there are also common idiomatic expressions in Brazilian Portuguese for words such as ``tiroteio” (``shootout”) that are unrelated to actual firearm shootings, as in ``mais perdido que cego em tiroteio” (literally, ``more lost than a blind person in a shootout”). 

Therefore, manually reviewing social media data is important to discern which messages are actual reports of gun violence. Analysts used Tweetdeck (now X Pro) to browse messages with keywords associated with gun violence. The analysts search for keywords associated with gun violence on Tweetdeck\footnote{\href{https://tweetdeck.twitter.com/}{https://tweetdeck.twitter.com/}} and filter the results by location, considerably narrowing down the search scope. Twitter documentation\footnote{\href{https://developer.twitter.com/en/docs/tutorials/advanced-filtering-for-geo-data}{https://developer.twitter.com/en/docs/tutorials/advanced-filtering-for-geo-data}} estimates that only 1-2\% of tweets messages are geo-tagged, but 30-40\% of them contain profile location information. During our intervention, Tweetdeck geographic search used both fields to filter messages from a specific location. Still, filtering messages to get only those with user or post-level location metadata ignores the majority of the data available on Twitter, and searching without the geographical filter would lead to an overload of unrelated information. 

Once a citizen report of gun violence is identified, the analysts interact with the user, sending a semi-standardized message to request further details about the time and location of the event. Below, we define this type of reply as an \textbf{interaction} from Fogo Cruzado with the users. 

Importantly, not all reports receive an interaction from Fogo Cruzado's team. Reports of gun violence using the keywords monitored and with location metadata associated might not receive an interaction due to several reasons. For example, the analyst may not see a particular message, or the event may have already been registered. For security reasons, analysts also refrain from engaging with users who display any indications of association with criminal activities. 

\section{Methodology}
\label{chap:methodology}
To evaluate a Transformer-based model's ability to detect gun violence reports in Portuguese on Twitter, we developed a prototype for real-time tweet classification and tested it with Fogo Cruzado analysts in Rio de Janeiro. Next, we describe our mixed method approach, integrating quantitative and qualitative analysis, to gain a thorough understanding of the challenges and impacts of AI models in the context of human rights monitoring.

\subsection{Data collection}
\label{subchap:datasource}
We collected labeled and unlabeled data using the full-archive search endpoint from the Twitter Academic API v2 between December 2022 and August 2023. 

First, we collected a \textbf{Dataset $L$} of labeled examples for the positive class, with tweets about gun violence reports). The dataset includes all posts that received a reply from Fogo Cruzado's profile in Rio de Janeiro requesting information about gun violence events. The labeled dataset $L$ includes all Twitter posts to which the user ``\nolinkurl{@fogocruzadorj}'' responded to, totaling 36,241 messages posted between 9 June 2016 and 24 May 2023.

Then, we collected a \textbf{Dataset $U_{lang}$} of unlabeled examples for the negative class, with tweets in Portuguese with keywords monitored by Fogo Cruzado but not responded to. We included all posts with the same keywords monitored by Fogo Cruzado using distinct search parameters, regardless of the user location. We excluded all posts for which the language metadata is not Portuguese (``\textit{lang: pt}''). The dataset contains 12,803,338 messages posted between 5 December 2020 and 7 May 2023.

Finally, we collected a \textbf{Dataset $U_{geo}$} of unlabeled examples for the negative class, with posts in Rio de Janeiro with keywords monitored by Fogo Cruzado but not responded to. As before, we included the same keywords monitored by Fogo Cruzado. We excluded all posts for which the geolocation metadata is not in Rio de Janeiro. This dataset specifically emulates the Tweetdeck feed monitored by Fogo Cruzado and contains 319,114 messages posted between 31st October 2016 and 7th May 2023. Most of these messages are examples of the negative class (i.e., do not represent a report of gun violence event). However, as explained in Section~\ref{chap:background}, there are also reports that have not received an interaction and, hence, are not labeled. Assuming that gun violence is more common in Rio de Janeiro than in other Portuguese-speaking regions, posts with geographical metadata associated with Rio de Janeiro ($ U_{geo}$) are more likely to have a higher proportion of reports of gun violence compared to all posts in Portuguese containing the keywords monitored ($ U_{lang}$). 

We filtered $L$, $U_{lang}$, and $U_{geo}$ with the following preprocessing rules: all tweets with mentions to the partner organization username were removed, mentions and links were replaced with special tokens, extra spaces were trimmed, and we deleted all tweet replies and duplicated messages, ignoring unavailable tweets. 

Because emojis encode important semantic information~\citep{kirk-etal-2022-hatemoji} and our base model~\citep{souzaBERTimbauPretrainedBERT2020} lacks representations for them, we opted to convert emojis into text descriptions. For instance, the music notes emojis might denote that the text is quoting song lyrics. Converting emojis to text provides a good tradeoff between fully retraining the model and ignoring emojis, hence losing important information, as the model would represent them as a single unknown token.

Table~\ref{table:samples} shows examples of both classes in our training dataset. Most of the messages in the training dataset are relatively short. The average number of words per message in the training dataset is 16.

\begin{table}[ht]
  \centering
  \caption{Examples of the positive and negative classes (authors' translations to English).}
  \label{table:samples}
  \begin{tabular}{p{0.45\textwidth}p{0.45\textwidth}}
    \toprule
    \textbf{Positive Examples ($L$)} & \textbf{Negative Examples ($U_{lang}$, $U_{geo}$)}\\
    \midrule
    Gunshots started going off right when I ordered a milkshake. I hope this man is a brave warrior. & Sometimes, certain words are like a shot, especially when you're feeling a little insecure \\  
    \midrule
    People here randomly fire off shots out of nowhere. & Oh so many distracted friends' photos that I took Jesus I deserved to get shot lol \\
    \midrule
    I already wake up startled, hearing gunshots. & I'm trying to let my nails grow, but when anxiety attacks, I tear them all off.\\
    \bottomrule
  \end{tabular}
  \vspace{0.5cm}
\end{table}

\subsection{Model development}
\label{subsec:model-development}
We implemented a dummy classifier, a Naive-Bayes, and a Transformer model. The dummy classifier is a random baseline that ignores text and generates stratified random predictions. We used the Naive-Bayes model with a TF-IDF vectoriser as a secondary baseline.

To implement the Transformer model, we leveraged BERTimbau\footnote{\href{https://huggingface.co/neuralmind/bert-large-portuguese-cased}{https://huggingface.co/neuralmind/bert-large-portuguese-cased}}, a pre-trained BERT model with Portuguese text~\citep{souzaBERTimbauPretrainedBERT2020}. BERTimbau was trained using \textit{brWaC}, a corpus with 3.5 million web pages in Brazilian Portuguese, processed through a model with 24 layers, 1024 hidden dimensions, 16 attention heads and 330 million parameters~\citep{wagnerfilhoBrWaCCorpusNew2018,souzaBERTimbauPretrainedBERT2020}. We fine-tuned the model for binary classification. This process involves incorporating an additional output layer into the neural network, which represents the probabilities associated with the positive and negative classes. We employed the Adam optimizer with a learning rate of $2.10^{-5}$. To prevent overfitting, we applied a dropout probability of 5\% to the hidden and attention layers and implemented early stopping to terminate the training process. For computational efficiency, we reduced the maximum token length value (from 512 to 128) after analyzing the token length distribution in the training data. 

We adopted self-training to address the challenge of making inferences from partially labeled data. Self-training is a long-standing semi-supervised learning method~\citep{aminiSelfTrainingSurvey2022}. It initially trains the model with a reduced number of labeled data and then uses it to generate pseudo labels for unlabeled data. These pseudo-labels are then combined with the original labeled data, and the augmented dataset is used to retrain the model to enhance its performance, a process that can be repeated one or more times in a loop.  

We trained the models on data prior to January 2023 and reserved data between January and May 2023 for holdout datasets. Compared to a random allocation between training and holdout datasets, this temporal split allows us to test the model on the most recent data. The approach seeks to approximate the real-world scenario analysts would encounter when monitoring never-seen data, with distributional shift (new location names or events). This criterion yielded 24,353 messages in \textbf{Dataset \textit{L}} that were used as labeled examples of the positive class to train the model. We defined the negative class filtering \textbf{Dataset $U_{lang}$} by messages posted before 2023 without location metadata. Then, we deleted those posts overlapping the \textbf{Dataset \textit{L}} and down-sampled the data, randomly selecting a volume of messages three times the number of positive samples (73,059) to avoid an extreme class imbalance. Therefore, the initial training dataset combining both classes had 97,412 messages.

After fine-tuning the Transformer model, we used it to generate pseudo-labels for the messages in \textbf{Dataset $U_{geo}$} posted before 2023 and not used in the initial training. This self-training approach resulted in 199,015 labeled examples used to augment the training dataset. To validate the pseudolabels generated, we categorized the predictions into four quantiles according to the probability assigned to each positive class. So, we selected ten samples from each class and quantile, resulting in a validation sample of 80 messages. Finally, we augmented the original training dataset to include the pseudo labels and fine-tuned BERTimbau again from scratch.

\begin{figure}[htb]
    \centering
    \includegraphics[width=0.7\linewidth]{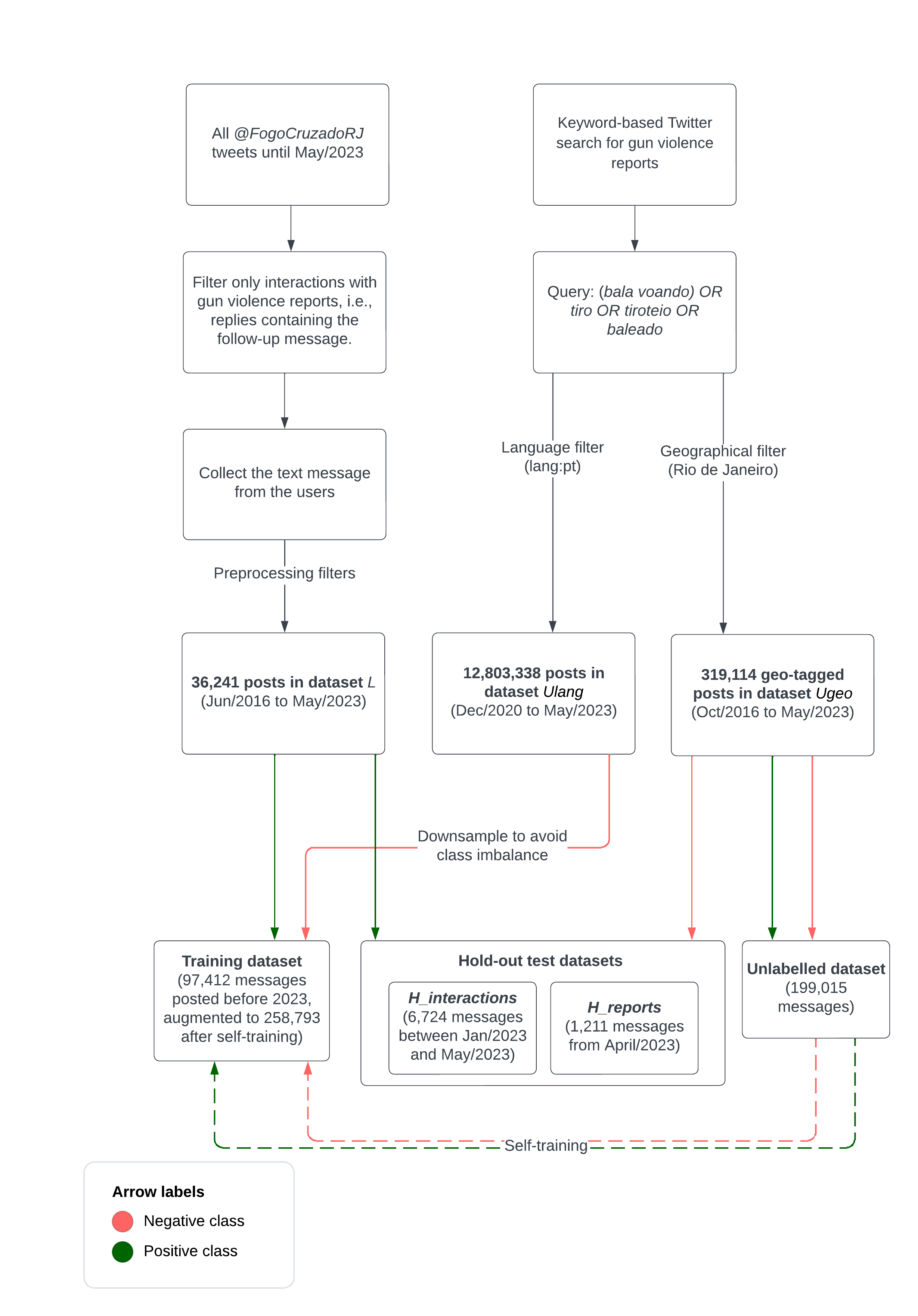}
    \caption{Data collection and transformations to create training and testing datasets.}
    \label{app:datasets}
\end{figure}

Figure~\ref{app:datasets} summarizes the data collection and transformation process to train and test the model. Green and red arrows represent, respectively, positive and negative classes of texts. 

\subsection{Model evaluation}
To evaluate the model's ability to accurately identify gun violence reports from social media posts, we report the model performance using precision, recall, and F1-score of the positive class. We prioritize recall over other evaluation metrics because, in our context, the presence of false negatives poses a greater risk than false positives. False negative cases imply potential oversight of gun violence reports due to model misclassification. In contrast, false positives only introduce additional labor with unrelated messages presented for human review. 

We created two holdout datasets by filtering both datasets $L$ and $ U_{geo} $ (see Section~\ref{subchap:datasource}). \textbf{Dataset $ H_{interactions} $} aims to measure the model performance to predict Twitter interactions done by Fogo Cruzado's analysts. In addition, \textbf{Dataset $ H_{reports} $} aims to measure the overall ability to distinguish between gun violence reports and ordinary texts.  

\textbf{Dataset $ H_{interactions} $} contains 6,724 messages from 2023 or later. The positive class has 1,909 posts resulting from filtering \textbf{Dataset \textit{L}} for posts dated 1st January 2023 or later. The negative class has 4,815 unlabeled posts from \textbf{Dataset $ U_{geo} $}. We excluded messages posted before 2023 and those already in the \textbf{Dataset $L$}. Some of the messages assigned to the negative class are actual gun violence reports that never received an interaction with Fogo Cruzado, potentially increasing false positive error rates. Yet, we are interested specifically in the recall, which disregards false positive cases.

\textbf{Dataset $ H_{reports} $} is a manually validated holdout subset of \textbf{Dataset ($ H_{interactions} $)} with 1,211 posts dated April 2023. These posts have all been coded by the first author, a native Brazilian Portuguese speaker, who has manually changed the labels of 147 (12\%) messages containing gun violence reports that were initially labeled as negative cases for having not received an interaction from Fogo Cruzado.

Finally, we further conducted a blind manual validation of a subset of 300 predictions from \textbf{Dataset $ H_{interactions}$} to investigate model reliability. The first author assigned labels to 150 random records from each class. We then compared these labels to the predictions of the best-performing model. This manual validation allowed us to compare these human-assigned labels with the best-performing model's predictions in a blind manner, meaning the author did not have access to the model’s predictions while labeling. 

\subsection{Designing a real-world intervention}
\label{subchap:intervention}
We designed an intervention with the Fogo Cruzado team to investigate the model performance in a real-world setting. We created a prototype, and Fogo Cruzado has adopted it as part of the standard workflow for live monitoring of new reports of gun violence in Rio de Janeiro. 

The prototype used the Twitter API v1 (search endpoint) to retrieve the latest tweets and a (CPU-only) server for preprocessing and classifying them using the best-performing model. We developed a Python script to execute this pipeline and upload the results to a web platform. Based on insights gained from the preliminary interviews with Fogo Cruzado's team leader in Rio de Janeiro, the prototype was initially configured to run every 15 minutes; later, we reduced this interval to five minutes at the participants’ request.

Fogo Cruzado's team in Rio de Janeiro adopted our AI-powered prototype to browse tweets from 29 May to 2 July 2023. Five professionals engaged directly with this intervention. 

Our prototype presents live Twitter data aggregated by different tabs. Messages classified as potential reports in Rio de Janeiro and those classified as negative cases are aggregated in the first two tabs. The third tab displays positive cases of messages without location metadata or users' location descriptions not matched to Rio de Janeiro. The information is displayed in tables: each row represents a message, and the columns are the text, timestamp, user location and profile bio. The interface allowed analysts to click on a button to view the original post on Twitter or respond to the user with a standardized follow-up message. 

\begin{figure}[h]
    \centering
    \begin{subfigure}[t]{0.45\textwidth}
        \centering
        \vspace{0pt} 
        \includegraphics[width=\textwidth]{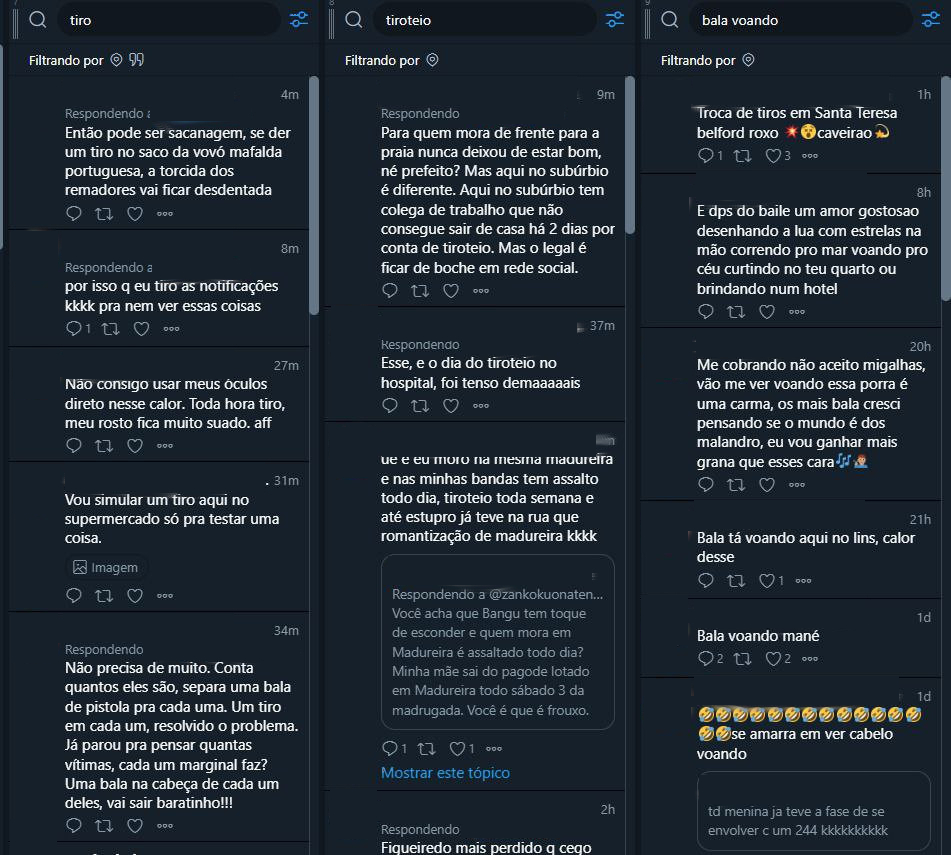}
        \caption{Column-based Tweetdeck layout for analysts to search relevant posts.}
        \label{fig:tweetdeck}
    \end{subfigure}
    \hfill
    \begin{subfigure}[t]{0.45\textwidth}
        \centering
        \vspace{0pt} 
        \includegraphics[width=\textwidth]{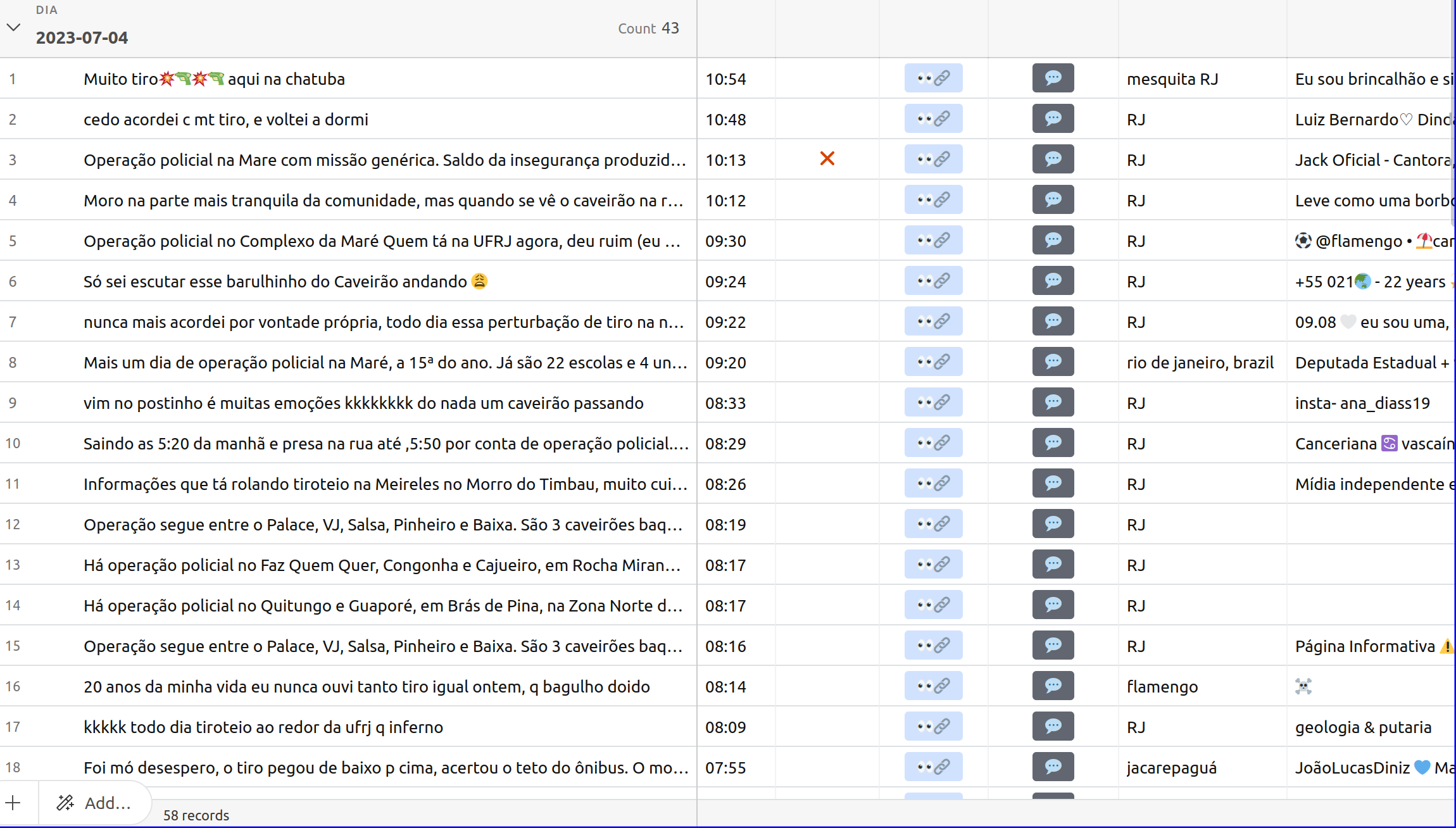}
        \caption{Our intervention prototype for analysts to review relevant posts identified by our model.}
        \label{fig:airtable}
    \end{subfigure}
    \caption{Interface screenshots for the Tweetdeck timelines (a) and intervention webapp (b).}
    \label{fig:screenshots}
\end{figure}

The prototype was expected to provide a more straightforward interface to identify gun violence reports on Twitter and allow analysts to interact with users. Figure~\ref{fig:tweetdeck} shows the Tweetdeck interface used by analysts to review posts. The column-based layout of Tweetdeck requires analysts to continuously scroll and click on the profiles to find new messages and the associated metadata. Figure~\ref{fig:airtable} shows our interface, which displays classified tweets and metadata in a structured table, allowing analysts to easily find the information they need for fact-checking.

We collected Fogo Cruzado's interaction on Twitter during the intervention and gathered information from the participants using surveys and interviews. We conducted all interviews and survey applications in May and July 2023.

Participants were asked to complete an online survey with closed-ended questions (Appendix~\ref{app:results}) before and after the intervention. The first survey gathered information to aid the interpretation of the training data and the standard workflow for online report monitoring. The post-intervention survey evaluated the impact of the model's adoption for real-time monitoring and gauged the subjective assessments of the participants. 

We conducted five interviews. We interviewed the analyst team leader in Rio de Janeiro before the intervention. The team leader and three analysts were also interviewed afterwards, approximately a month after the start of our intervention. The interviews followed a semi-structured approach, and we did thematic and content analysis~\citep{rossmanIntroductionQualitativeResearch2017} of the answers, aiming to identify recurring patterns of topics related to the RQ2.

\section{Results}
\label{chap:results}
In this section, we present the results obtained to evaluate the model's performance using the evaluation datasets (RQ1) and assess both the potential and shortcomings of its application for real-time monitoring (RQ2). Beyond quantitative metrics, such as accuracy and F1 scores, we emphasize the model's ability to assist human rights analysts in filtering relevant content from the vast amount of social media data. Based on surveys and interviews conducted during our intervention, we find that BERTimbau's performance in detecting gun violence reports significantly improved the efficiency of monitoring efforts.

\subsection{Model performance}

Table~\ref{tab:scores-raw} shows the precision, recall, and F1-score metrics for holdout \textbf{Dataset $ H_{interactions} $}, which measure the models' ability to correctly capture how Fogo Cruzado interact with Twitter users. BERTimbau reaches 91\% in recall of the positive class, meaning that it correctly predicted nine out of ten Fogo Cruzado's interactions between January and May 7th 2023. The manual validation for the predictions based on \textbf{Dataset $ H_{interactions} $} reveals a strong 95\% agreement rate between the labels assigned by BERTimbau and ours (285 agreements out of 300 posts manually analyzed).

\begin{table}[h]
\centering
\caption{Evaluation metrics for the holdout dataset $ H_{interactions}$, measuring if models capture analysts' interactions on Twitter}
\label{tab:scores-raw}
\begin{tabular}{lcccc}
\toprule
\textbf{Model} & \textbf{Precision} & \textbf{Recall} & \textbf{F1-score} \\
\midrule
Random baseline & 0.29 & 0.25 & 0.27 \\
Naive-Bayes & 0.70 & 0.83 & 0.76 \\
BERTimbau & \textbf{0.71} & \textbf{0.91} & \textbf{0.80} \\
\bottomrule
\end{tabular}
\end{table}

Table~\ref{tab:scores-label} shows  the precision, recall, and F1-score metrics for holdout \textbf{Dataset $ H_{reports}$}, which measure the overall performance of the models to distinguish ordinary Portuguese text from gun violence reports---regardless of Fogo Cruzado's interactions. BERTImbau reaches an F1-score of 90\% for the positive class. This suggests that our prototype powered by BERTImbau should be able to find a wider range of gun violence reports than the small set of posts captured by Fogo Cruzado's partial Tweetdeck filters. 

\begin{table}[h]
\centering
\caption{Evaluation metrics for the human-reviewed holdout dataset $ H_{reports}$, measuring if models can distinguish gun violence reports from unrelated posts}
\label{tab:scores-label}
\begin{tabular}{lcccc}
\toprule
\textbf{Model} & \textbf{Precision} & \textbf{Recall} & \textbf{F1-score} \\
\midrule
Random baseline & 0.33 & 0.19 & 0.24 \\
Naive-Bayes & \textbf{0.91} & 0.78 & 0.84 \\
BERTimbau  & \textbf{0.94} & \textbf{0.87} & \textbf{0.90} \\
\bottomrule
\end{tabular}
\end{table}

The self-training approach proved useful. Retraining the model with pseudo-labels improved BERTimbau's performance by two percentage points in recall for the positive class and a one percentage point improvement in F1-score in the human-reviewed (\textbf{Dataset $ H_{reports}$}). Appendix~\ref{app:model_evaluation} provides confusion matrix and ROC curve that illustrates the performance of the best-performing model (BERTimbau) after the self-training approach.

\subsection{Error analysis}
Most of the incorrect predictions of BERTimbau for \textbf{Dataset $ H_{interactions}$} are false positives (697 false positives out of 877 incorrect predictions), i.e. messages from the negative class classified by the model as gun violence reports (see Appendix~\ref{app:model_evaluation}). This result is expected because, as explained in Section \ref{chap:methodology}, there are actual reports of gun violence that have not received an interaction from Fogo Cruzado's team and, therefore were not assigned to the positive label in this evaluation dataset. On the contrary, the human-reviewed \textbf{Dataset $ H_{reports}$} had more messages misclassified as false negatives than false positives (61 false negatives out of 89 incorrect predictions).

To better understand the misclassifications, we analyzed linguistic features of the 89 misclassified instances and found two relevant patterns. The error analysis of \textbf{Dataset $ H_{reports}$} revealed that the fine-tuned BERTimbau model struggles with long text and emojis. First, the average length of the incorrect predictions (104 characters) is higher than the average length of the training dataset (80 characters). Secondly, emojis are present in 23\% of the incorrect predictions (compared to 13\% in the whole training dataset). 

\subsection{Intervention findings}
\label{subchap:results-intervention}

From 28 May to 2 July 2023, the prototype gathered 21,871 messages from Twitter, classified them, and automatically uploaded them to the web interface. To assess how effective the model and prototype were in a real-world monitoring application, we then conducted two rounds of online surveys and interviews with human rights analysts who used the prototype.

The survey suggests that Twitter was a crucial source of new evidence at the time. Participants noted that, prior to our intervention, the signal-to-noise ratio varied depending on the keywords used for the search, as some terms were frequently used in unrelated contexts. Reviewing Tweetdeck required a significant amount of time. Importantly, participants reported that verifying identified reports was even more time-consuming than the initial data collection phase, indicating that fully automating event detection in human rights research without expert supervision is currently unlikely to be reliable in practical applications. 

Still, survey participants agreed that the prototype improved their efficiency, allowing them to identify reports of gun violence more quickly and allocate their time more effectively. Although adoption of the tool was optional, all four participants reported daily use of the prototype. They evaluated the model performance as ``good'' and ``excellent.'' and estimated that more than 60\% of the reports identified by the model were validated and contributed to new records in the Fogo Cruzado database. Appendix~\ref{app:results} presents a detailed account of the responses to our survey.

The interviews confirmed that our model can accurately identify reports of gun violence in Brazilian Portuguese and is helpful for real-time monitoring. The participants confirmed that our prototype allowed them to track more messages without being burdened by unrelated information. Overall, they consider that the advantages in terms of efficiency outweigh the problems encountered, and the prototype has improved their efficiency and helped them save time. Fogo Cruzado's analysts have highlighted the classifier performance to accurately distinguish between reports of gun violence and ordinary messages posted on Twitter. All participants expressed initial discomfort with the new workflow and interface for monitoring tweets. However, interviews indicate they quickly recognized the prototype's value and adopted it in their daily work. 

The primary advantages reported in the interviews and indicated in the survey are outlined as follows:

\begin{itemize}
    \item \textbf{The prototype filters out less relevant social media content and provides a higher signal-to-noise input for analysts. This enables them to improve the monitoring and tracking of human rights events.} Reviewing messages on Twitter and searching for potential reports of gun violence became more efficient with the prototype. Analysts claimed that the interface increases their agility and helps to streamline their workflow, enabling them to allocate their time more effectively. This advantage can be illustrated by the following quote from an interview with one of the participants: ``[Now] I do not have to go hunting for tweets. Sometimes, I missed them [gun violence reports] because there were too many [unrelated] messages. During the BBB [Big Brother Brasil, an annual TV show extremely popular on Twitter], it was chaotic […]. It was literally a treasure hunt''\footnote{Authors' translation to English.}.

    \item \textbf{Instead of using restrictive geolocation filters that limit the number of messages when collecting data, our prototype allows analysts to expand their search scope.} The platform enabled users to search for more terms or review messages that were classified by the model as gun violence reports but do not contain location metadata, thus enabling them to scrutinize more reports. The prototype allowed analysts to analyze reports without being overwhelmed by the volume of messages. One of the participants deemed ``very important'' the tab showing all tweets with reports of gun violence, regardless of having location metadata associated with Rio de Janeiro. This participant noted that some of these tweets have user location information related to Rio de Janeiro that is not correctly associated with this region by the Twitter search: ``People write [location metadata] the way they want. Today I discovered a shooting because of this tab. They use details and slang [that are not filtered by the standard Twitter geolocation search]''.

    \item \textbf{Reviewing messages pre-classified by the language model reduces the cognitive effort required to find relevant information.} The table view used in our prototype was deemed more straight to the point and better organized when compared to the column-based interface of Tweetdeck. According to the analysts, the spreadsheet eliminates the need for constant scrolling and reduces their cognitive effort. One of the participants reported that reviewing messages on Tweetdeck ``gives extra mental fatigue from processing much information at the same time.” The prototype was deemed particularly useful when the analysts needed to retrospectively review a high volume of messages. The analyst in the morning shift, who is responsible for reviewing posts between midnight and 6 am, describes this advantage as follows: ``Once I got the hang of it, it has been a thousand times better to scan [reports] on the morning shift [because before] I had to scroll down all Tweetdeck columns.''

\end{itemize}

Conversely, the interviews and surveys also revealed limitations. The use of our prototype to gather information on events already identified, as opposed to discovering reports on new events or live track conflicts in real time, was limited. This happened mostly because of the following drawbacks, according to the participants:

\begin{itemize}
    \item \textbf{Updating new tweets promptly is critical for monitoring ongoing conflicts}: The fact that the prototype only updated the messages in the web interface in five-minute intervals limited the analysts' abilities to quickly monitor events as they happen. They also pointed out that a timestamp indicating the last data update should be included.

    \item \textbf{Search keywords need to be dynamically set}: Custom searches are part of analysts' standard workflow to monitor events in Tweetdeck. They add terms related to specific locations to verify and fact-check reports. However, the prototype did not allow users to change the keywords used to fetch data from Twitter. 

    \item \textbf{Further information beyond texts can help analysts interact with users}: Analysts highlighted the absence of user photos as a drawback. According to them, user photos are important to determine whether there will be a follow-up interaction requesting further information about the event. Using the prototype to monitor reports required an extra click to check the user profile and picture.

\end{itemize}

Overall, the participants have adopted the platform due to its advantages in terms of broader search capabilities, time optimization, and interface improvements. Participants reported that sometimes the model misclassified long texts or posts reproducing music lyrics. One of the participants also noted that posts with lyrics might be challenging even for humans, as classifying them requires contextual knowledge. The interviews revealed that the participants chose to use both the prototype and Tweetdeck concurrently instead of replacing one platform with the other. While our prototype was preferred to discover new cases, Tweetdeck was used mainly to observe events as they unfold and follow specific Twitter profiles.

The mean number of interactions from Fogo Cruzado's profile in Rio de Janeiro (\nolinkurl{@FogoCruzadoRJ}) with users reporting gun violence before the intervention was 17 and increased to 24 after adopting the prototype. To test if the intervention was indeed associated with the increases in interactions, we used a difference-in-difference analysis. We defined Rio de Janeiro as a treatment group and Fogo Cruzado's team in Bahia as a control group. We used the following control variables to ensure that the observed changes in interaction rates were not influenced by other external factors: the number of gun violence events, the total number of victims (killed or injured), and the population average of the affected cities. We used Fogo Cruzado's public database \footnote{\url{https://api.fogocruzado.org.br/}} to collect data on gun violence events and victims. Appendix~\ref{app:diffindiff} provides further details on the variables and the model used for our difference-in-difference analysis. 

The difference-in-difference regression shows that our intervention is associated with an increase in nine interactions with reports of gun violence per day. This result is aligned with the interviews and survey findings: all four analysts consulted stated that the model was useful or very useful in enabling them to identify more reports, as shown in Figure~\ref{fig:postsurvey3} (Section~\ref{app:questionnaire}).

\section{Discussion}
\label{chap:discussion}
Our work combines quantitative and qualitative methods to provide the first systematic assessment of adopting a language model utilizing social media data to assist human rights monitoring in real-world settings. Our evaluations of the models' performance confirm that Transformer models are suitable for classification tasks in quantitative human rights research and applications. 

BERTimbau strongly outperforms the random baseline but provides smaller improvements over Naive-Bayes for some metrics in Tables~\ref{tab:scores-raw} and \ref{tab:scores-label}. The Naive Bayes model relies on a naive assumption of feature independence---assuming words in a text are independent of each other---, suggesting that its relatively high precision can be due to the brevity of the messages and the strong predictive power of certain words in Twitter posts (see Section~\ref{chap:methodology}). For our intervention, BERTimbau significantly reduces false negative cases and provides a substantial performance gain for our main metric of interest, the recall score (0.78 for NB to 0.87 for BERTimbau in \textbf{Dataset $ H_{reports} $}, the manually-coded evaluation set), both justifying the choice of the Transformer model over a simpler model. The performance metrics are on par with or outperformed scores reported in similar previous studies~\citep{taGANBERTAdversarialLearning2022,alhelbawyNLPPoweredHumanRights2020,gokhaleDeployingCotrainingAlgorithm2017}. Although these studies are not directly comparable due to differences in methodologies and datasets, they collectively demonstrate that machine learning models can aid real-time human rights monitoring by filtering the signal from noise.

Fine-tuning BERTimbau with the self-training approach required approximately six GPU hours (GPU A100). For inference, we ran the model using existing infrastructure with CPUs only; therefore, there were no new costs after the training phase. This approach ensures scalability, as the trained model can be deployed even on basic web servers, allowing for efficient processing and relatively low latency (approximately one second to classify each message) without the need for additional GPU resources. On the other hand, unlike zero-shot classification approaches with generative AI, classification using BERT requires retraining the model to adapt it to other contexts.

Our intervention shows that language models can help human rights analysts filter social media posts in order to find evidence of human rights violations. It is worth highlighting that participating in our research and using the prototype were presented as optional alternatives, by no means mandatory or imposed by supervisors. With an already high number of sources to monitor, the analysts could simply choose not to use the prototype if they had not considered it useful. Nevertheless, the team adopted it as one of their standard tools, and Fogo Cruzado continued to employ our prototype for social media monitoring even after the research ended. 

In July 2023, Twitter revoked the Academic API access, but an alternative data collection method allowed the prototype to remain in use until March 2024, when restrictions were further tightened. The reliance on a single source of data is a critical limitation of our prototype. Changes in the company’s leadership have led to drastic restrictions in the API access policy. As of this writing, accessing more than 3,000 tweets via the API costs 5,000 US dollars per month, an unaffordable price for most NGOs, especially those in middle and low-income countries.

The long-term sustainability of social media monitoring systems by human rights organizations is challenging due to the high costs of adapting to the ever-changing technical and social context. In addition to data collection restrictions, user engagement across different platforms evolves over time, potentially demanding more investments to collect data from other sources. Similarly, in the long run, adapting the model to ever-changing linguistic dynamics might be necessary. For example, the expressions used to describe gun violence events five years from now may not be the same as those used today. Although techniques such as continuous training~\citep{baylorContinuousTrainingProduction2019} can be leveraged to refine the model’s performance continuously and improve accuracy over time, implementing and maintaining these solutions is often unaffordable for non-profits.

With limited financial resources, enhancing data capabilities in the human rights sector requires collaboration with other nonprofits and tech-savvy partners~\citep{farmerDataSocialGood2023}. Academics, in particular, could strengthen a mutually beneficial relationship with human rights organizations. We echo the call by Lazer et al. for academic researchers to prioritize real-world problems and enhance collaboration with non-academic actors~\citep{lazerComputationalSocialScience2020}; however, the authors fall short by listing only industry and government as potential partners. Non-governmental organizations have made the most cutting-edge advances in quantitative human rights studies~\citep{unitednationsHumanRightsIndicators2013,goodhartHumanRightsPolitics2016} and continue to collect on-the-ground information valuable for academic research. In turn, academics could contribute not only by scaling current data collections but also by working towards further public interest technologies in human rights studies.

Our intervention provides strong evidence that Transformer-based architectures can effectively support human rights analysts in filtering high volumes of data to identify reports of human rights violations with low error rates, aligning with prior research in this domain~\citep{nemkovaDetectingHumanRights2023,alhelbawyNLPPoweredHumanRights2020}. Despite achieving excellent results in qualitative and quantitative assessments, our model can occasionally miss or misclassify reports. Importantly, these errors may predominantly affect specific users or texts and lead to selection biases, a critical issue for quantitative human rights research~\citep{price_selection_2015}. The participants decided to use our prototype \emph{alongside} their existing information retrieval approaches, suggesting that automated models with expert oversight should be complementing traditional search methods, rather than replacing them. Using different systems in parallel can enhance accuracy and ensure nuanced contextual understanding. Automated models may also support other unmet needs, such as preserving digital evidence~\citep{mooneySocialMediaEvidence2021}; however, human intervention in maintaining curated and regularly updated keywords for search~\citep{myersHowConductDiscovery2020} remains an important requirement for their effectiveness.

Finally, our results show that current open-source AI models can support crowdsourcing initiatives using social media data for human rights monitoring. However, questions remain about how these models might influence analysts' approaches to documenting events. In the best scenario, AI can help mitigate the underreporting of cases. Alternatively, they might lead to over- or under-representation of certain groups based on linguistic features. They could also lead to the identification of more reports of the same events already observed without automation, preserving and reinforcing blind spots of data collection. It is unclear if and how adopting automated systems for social media monitoring contributes to statistical biases and the ``information effect”\cite{greeneMachineLearningHuman2019,clarkInformationEffectsHuman2013}, in which upward trends in event counts occur because human rights monitors became able to observe more cases. Overall, while AI holds promise for enhancing human rights monitoring, further research is needed to understand its impact and to develop strategies that ensure it effectively supports organizations in documenting events accurately and comprehensively.

\section{Conclusion}
\label{chap:conclusion}

As organizations incorporate machine learning models and other data-driven techniques to analyze human rights violations, addressing potential biases and fairness concerns is crucial. First, unequal access to resources in terms of knowledge and financial means can create disparities in the capacity to leverage machine learning and open-source information, potentially widening the gap between well-funded and under-resourced organizations. The financial and human resources needed to train open-source language models and keep them functional are bottlenecks for human rights organizations. Commercial language models might be a cheaper alternative, but they often lack transparency, induce users' dependency on a third-party service and are general-purpose solutions, as opposed to models tailored for human rights investigations. Second, biases related to collecting user-generated content may introduce or reinforce biases in data, as information from specific demographics or regions may be overrepresented or underrepresented, potentially leading to inaccurate conclusions. Finally, the unequal application of machine learning across different areas of human rights may result in some violations being better addressed than others, depending on data availability or the suitability of this technique. Encouraging collaboration between different actors, developing guidelines and benchmarks, as well as promoting capacity-building initiatives can help ensure that these modern techniques are employed in a manner that is both equitable and effective.

\bibliographystyle{plainnat}
\bibliography{bibliography}

\begin{thebibliography}{61}
\providecommand{\natexlab}[1]{#1}
\providecommand{\url}[1]{\texttt{#1}}
\expandafter\ifx\csname urlstyle\endcsname\relax
  \providecommand{\doi}[1]{doi: #1}\else
  \providecommand{\doi}{doi: \begingroup \urlstyle{rm}\Url}\fi

\bibitem[Abbas et~al.(2021)Abbas, Pitt, and
  Michael]{abbasSocioTechnicalDesignPublic2021}
Roba Abbas, Jeremy Pitt, and Katina Michael.
\newblock Socio-{{Technical Design}} for {{Public Interest Technology}}.
\newblock 2\penalty0 (2):\penalty0 55--61, 2021.
\newblock ISSN 2637-6415.
\newblock \doi{10.1109/TTS.2021.3086260}.
\newblock URL \url{https://ieeexplore.ieee.org/abstract/document/9459499}.

\bibitem[Alhelbawy et~al.(2016)Alhelbawy, Massimo, and
  Kruschwitz]{alhelbawyCorpusViolenceActs2016}
Ayman Alhelbawy, Poesio Massimo, and Udo Kruschwitz.
\newblock Towards a {Corpus} of {Violence} {Acts} in {Arabic} {Social} {Media}.
\newblock In \emph{Proceedings of the {Tenth} {International} {Conference} on
  {Language} {Resources} and {Evaluation} ({LREC}'16)}, pages 1627--1631.
  European Language Resources Association, 2016.
\newblock URL \url{https://aclanthology.org/L16-1257}.

\bibitem[Alhelbawy et~al.(2020)Alhelbawy, Lattimer, Kruschwitz, Fox, and
  Poesio]{alhelbawyNLPPoweredHumanRights2020}
Ayman Alhelbawy, Mark Lattimer, Udo Kruschwitz, Chris Fox, and Massimo Poesio.
\newblock An {NLP}-{Powered} {Human} {Rights} {Monitoring} {Platform}.
\newblock \emph{Expert Systems with Applications}, 153, 2020.
\newblock ISSN 0957-4174.
\newblock URL \url{https://doi.org/10.1016/j.eswa.2020.113365}.

\bibitem[Amini et~al.(2022)Amini, Feofanov, Pauletto, Devijver, and
  Maximov]{aminiSelfTrainingSurvey2022}
Massih-Reza Amini, Vasilii Feofanov, Loic Pauletto, Emilie Devijver, and Yury
  Maximov.
\newblock Self-{Training}: {A} {Survey}, 2022.
\newblock URL \url{https://doi.org/10.48550/arXiv.2202.12040}.

\bibitem[Arellano et~al.(2022)Arellano, Escalante, Villaseñor~Pineda, Montes~y
  Gómez, and Sanchez-Vega]{arellanoOverviewDAVINCISIberLEF2022}
Luis~Joaquín Arellano, Hugo~Jair Escalante, Luis Villaseñor~Pineda, Manuel
  Montes~y Gómez, and Fernando Sanchez-Vega.
\newblock Overview of {DA}-{VINCIS} at {IberLEF} 2022: {Detection} of
  {Aggressive} and {Violent} {Incidents} from {Social} {Media} in {Spanish}.
\newblock \emph{Procesamiento del Lenguaje Natural}, 69, 2022.
\newblock ISSN 1135-5948.
\newblock URL \url{https://doi.org/10.26342/2022-69-18}.

\bibitem[Ball and Price(2019)]{ball_using_2019}
Patrick Ball and Megan Price.
\newblock Using statistics to assess lethal violence in civil and inter-state
  war.
\newblock \emph{Annual review of statistics and its application}, 6:\penalty0
  63--84, 2019.
\newblock URL \url{https://doi.org/10.1146/annurev-statistics-030718-105222}.

\bibitem[Bauer et~al.(2022)Bauer, Longley, Ma, and
  Wilson]{bauerNLPHumanRights2022}
Daniel Bauer, Tom Longley, Yueen Ma, and Tony Wilson.
\newblock {NLP} in {Human} {Rights} {Research}: {Extracting} {Knowledge}
  {Graphs} about {Police} and {Army} {Units} and {Their} {Commanders}.
\newblock In \emph{Proceedings of the 16th {Linguistic} {Annotation} {Workshop}
  ({LAW}-{XVI}) within {LREC2022}}, pages 62--69. European Language Resources
  Association, 2022.
\newblock URL \url{https://aclanthology.org/2022.law-1.7}.

\bibitem[Baylor et~al.(2019)Baylor, Haas, Katsiapis, Leong, Liu, Menwald, Miao,
  Polyzotis, Trott, and Zinkevich]{baylorContinuousTrainingProduction2019}
Denis Baylor, Kevin Haas, Konstantinos Katsiapis, Sammy Leong, Rose Liu,
  Clemens Menwald, Hui Miao, Neoklis Polyzotis, Mitchell Trott, and Martin
  Zinkevich.
\newblock Continuous {{Training}} for {{Production ML}} in the {{TensorFlow
  Extended}} ({{TFX}}) {{Platform}}.
\newblock pages 51--53, 2019.
\newblock ISBN 978-1-939133-00-7.
\newblock URL
  \url{https://www.usenix.org/conference/opml19/presentation/baylor}.

\bibitem[Bloem and Salemi(2021)]{bloem_covid-19_2021}
Jeffrey~R. Bloem and Colette Salemi.
\newblock {COVID}-19 and conflict.
\newblock \emph{World Development}, 140:\penalty0 105294, 2021.
\newblock ISSN 0305-750X.
\newblock URL \url{https://doi.org/10.1016/j.worlddev.2020.105294}.

\bibitem[Buenos and Lima(2023)]{buenos_anuario_2023}
Samira Buenos and Renato~Sérgio Lima.
\newblock Anuário {Brasileiro} de {Segurança} {Pública} 2023, 2023.
\newblock URL
  \url{https://forumseguranca.org.br/anuario-brasileiro-seguranca-publica/}.

\bibitem[Cavalcanti(2018)]{cavalcantiDinamicasViolenciaUrbana2017}
Ricardo~Caldas Cavalcanti.
\newblock As dinâmicas da violência urbana na américa latina.
\newblock 7\penalty0 (2):\penalty0 226--251, 2018.
\newblock ISSN 2236-6725.
\newblock \doi{10.5902/2236672531915}.
\newblock URL \url{https://periodicos.ufsm.br/seculoxxi/article/view/31915}.

\bibitem[Clark and Sikkink(2013)]{clarkInformationEffectsHuman2013}
Ann~Marie Clark and Kathryn Sikkink.
\newblock Information {Effects} and {Human} {Rights} {Data}: {Is} the {Good}
  {News} {About} {Increased} {Human} {Rights} {Information} {Bad} {News} for
  {Human} {Rights} {Measures}?
\newblock \emph{Human Rights Quarterly}, 35\penalty0 (3):\penalty0 539--568,
  2013.
\newblock ISSN 0275-0392.
\newblock URL \url{https://www.jstor.org/stable/24518073}.

\bibitem[Devlin et~al.(2018)Devlin, Chang, Lee, and
  Toutanova]{devlinBERTPretrainingDeep2018}
Jacob Devlin, Ming-Wei Chang, Kenton Lee, and Kristina Toutanova.
\newblock {BERT}: {Pre}-training of {Deep} {Bidirectional} {Transformers} for
  {Language} {Understanding}, 2018.
\newblock URL \url{https://doi.org/10.48550/arXiv.1810.04805}.

\bibitem[Eze(2023)]{ezeImpactGunViolence2023}
Anthony~Nnaemeka Eze.
\newblock Impact of gun violence.
\newblock 8\penalty0 (1), 2023.
\newblock ISSN 2397-5776.
\newblock \doi{10.1136/tsaco-2023-001314}.
\newblock URL \url{https://www.ncbi.nlm.nih.gov/pmc/articles/PMC10729124/}.

\bibitem[Farmer et~al.(2023)Farmer, McCosker, Albury, and
  Aryani]{farmerDataSocialGood2023}
Jane Farmer, Anthony McCosker, Kath Albury, and Amir Aryani.
\newblock \emph{Data for {{Social Good}}: {{Non-Profit Sector Data Projects}}}.
\newblock Springer Nature, 2023.
\newblock ISBN 978-981-19555-4-9.
\newblock \doi{10.1007/978-981-19-5554-9}.
\newblock URL \url{https://library.oapen.org/handle/20.500.12657/61326}.

\bibitem[Fontanarosa and
  Bibbins-Domingo(2022)]{fontanarosaUnrelentingEpidemicFirearm2022}
Phil~B. Fontanarosa and Kirsten Bibbins-Domingo.
\newblock The {{Unrelenting Epidemic}} of {{Firearm Violence}}.
\newblock 328\penalty0 (12):\penalty0 1201--1203, 2022.
\newblock ISSN 0098-7484.
\newblock \doi{10.1001/jama.2022.17293}.
\newblock URL \url{https://doi.org/10.1001/jama.2022.17293}.

\bibitem[Garbarino et~al.(2002)Garbarino, Bradshaw, and
  Vorrasi]{garbarinoMitigatingEffectsGun2002}
James Garbarino, Catherine~P. Bradshaw, and Joseph~A. Vorrasi.
\newblock Mitigating the {Effects} of {Gun} {Violence} on {Children} and
  {Youth}.
\newblock \emph{The Future of Children}, 12\penalty0 (2):\penalty0 72, 2002.
\newblock ISSN 10548289.
\newblock URL \url{https://doi.org/10.2307/1602739}.

\bibitem[General(2024)]{generalFirearmViolenceAmerica2024}
Office of the~Surgeon General.
\newblock Firearm {{Violence}} in {{America}}, 2024.
\newblock URL
  \url{https://www.hhs.gov/surgeongeneral/reports-and-publications/firearm-violence/index.html}.

\bibitem[Gokhale and Fasli(2017)]{gokhaleDeployingCotrainingAlgorithm2017}
Ragini Gokhale and Maria Fasli.
\newblock Deploying a co-training algorithm to classify human-rights abuses.
\newblock In \emph{2017 {International} {Conference} on the {Frontiers} and
  {Advances} in {Data} {Science} ({FADS})}, pages 108--113, 2017.
\newblock URL \url{https://doi.org/10.1109/FADS.2017.8253206}.

\bibitem[Goodhart(2016)]{goodhartHumanRightsPolitics2016}
Michael Goodhart.
\newblock \emph{Human {Rights}: {Politics} and {Practice}}.
\newblock Oxford University Press, 2016.
\newblock ISBN 978-0-19-870876-6.
\newblock URL \url{https://doi.org/10.1093/hepl/9780198708766.001.0001}.

\bibitem[Greene et~al.(2019)Greene, Park, and
  Colaresi]{greeneMachineLearningHuman2019}
Kevin~T. Greene, Baekkwan Park, and Michael Colaresi.
\newblock Machine {Learning} {Human} {Rights} and {Wrongs}: {How} the
  {Successes} and {Failures} of {Supervised} {Learning} {Algorithms} {Can}
  {Inform} the {Debate} {About} {Information} {Effects}.
\newblock \emph{Political Analysis}, 27\penalty0 (2):\penalty0 223--230, 2019.
\newblock ISSN 1047-1987, 1476-4989.
\newblock URL \url{https://doi.org/10.1017/pan.2018.11}.

\bibitem[Hirata and Couto(2022)]{hirataMapaHistoricoDos2022}
Daniel Hirata and Maria~Isabel Couto.
\newblock Mapa {Histórico} dos {Grupos} {Armados} no {Rio} de {Janeiro}, 2022.
\newblock URL
  \url{https://geni.uff.br/2022/09/13/mapa-historico-dos-grupos-armados-no-rio-de-janeiro/}.

\bibitem[Hirata and Grillo(2019)]{hirataRoubosProtecaoPatrimonial2019}
Daniel Hirata and Carolina~Christoph Grillo.
\newblock Roubos, proteção patrimonial e letalidade no {Rio} de {Janeiro},
  2019.
\newblock URL
  \url{https://geni.uff.br/2021/03/26/roubos-protecao-patrimonial-e-letalidade-no-rio-de-janeiro/}.

\bibitem[Hirata et~al.(2023)Hirata, Grillo, Dirk, and
  Lyra]{hirataChacinasPoliciaisNo2023}
Daniel Hirata, Carolina~Christoph Grillo, Renata~Coelho Dirk, and Diego~Azevedo
  Lyra.
\newblock Chacinas {Policiais} no {Rio} de {Janeiro}: {Estatização} das
  mortes, mega chacinas policiais e impunidade, 2023.
\newblock URL
  \url{https://geni.uff.br/2023/05/05/chacinas-policiais-no-rio-de-janeiro-estatizacao-das-mortes-mega-chacinas-policiais-e-impunidade/}.

\bibitem[Hu et~al.(2022)Hu, Hosseini, Skorupa~Parolin, Osorio, Khan, Brandt,
  and D’Orazio]{hu_conflibert_2022}
Yibo Hu, MohammadSaleh Hosseini, Erick Skorupa~Parolin, Javier Osorio, Latifur
  Khan, Patrick Brandt, and Vito D’Orazio.
\newblock {ConfliBERT}: {A} {Pre}-trained {Language} {Model} for {Political}
  {Conflict} and {Violence}.
\newblock In \emph{Proceedings of the 2022 {Conference} of the {North}
  {American} {Chapter} of the {Association} for {Computational} {Linguistics}:
  {Human} {Language} {Technologies}}, pages 5469--5482. Association for
  Computational Linguistics, 2022.
\newblock URL \url{https://doi.org/10.18653/v1/2022.naacl-main.400}.

\bibitem[International(2011)]{amnesty_international_amnesty_2011}
Amnesty International.
\newblock The amnesty international timeline, 2011.
\newblock URL
  \url{http://static.amnesty.org/ai50/ai50-amnesty-international-timeline.pdf}.

\bibitem[Jabine and Claude(1992)]{jabine_human_1992}
Thomas~B. Jabine and Richard~P. Claude, editors.
\newblock \emph{Human {Rights} and {Statistics}: {Getting} the {Record}
  {Straight}}.
\newblock University of Pennsylvania Press, 1992.
\newblock ISBN 978-1-5128-0286-3.

\bibitem[Kalesan et~al.(2017)Kalesan, Adhikarla, Pressley, Fagan, Xuan, Siegel,
  and Galea]{kalesanHiddenEpidemicFirearm2017}
Bindu Kalesan, Chandana Adhikarla, Joyce~C. Pressley, Jeffrey~A. Fagan, Ziming
  Xuan, Michael~B. Siegel, and Sandro Galea.
\newblock The {{Hidden Epidemic}} of {{Firearm Injury}}: {{Increasing Firearm
  Injury Rates During}} 2001–2013.
\newblock 185\penalty0 (7):\penalty0 546--553, 2017.
\newblock ISSN 0002-9262.
\newblock \doi{10.1093/aje/kww147}.
\newblock URL \url{https://doi.org/10.1093/aje/kww147}.

\bibitem[Kirk et~al.(2022)Kirk, Vidgen, Rottger, Thrush, and
  Hale]{kirk-etal-2022-hatemoji}
Hannah Kirk, Bertie Vidgen, Paul Rottger, Tristan Thrush, and Scott Hale.
\newblock {H}atemoji: A test suite and adversarially-generated dataset for
  benchmarking and detecting emoji-based hate.
\newblock In Marine Carpuat, Marie-Catherine de~Marneffe, and Ivan~Vladimir
  Meza~Ruiz, editors, \emph{Proceedings of the 2022 Conference of the North
  American Chapter of the Association for Computational Linguistics: Human
  Language Technologies}, pages 1352--1368, Seattle, United States, July 2022.
  Association for Computational Linguistics.
\newblock \doi{10.18653/v1/2022.naacl-main.97}.
\newblock URL \url{https://aclanthology.org/2022.naacl-main.97}.

\bibitem[Koenig and Freeman(2020)]{koenigOpenSourceInvestigations2020}
Alexa Koenig and Lindsay Freeman.
\newblock \emph{Open {{Source Investigations}} for {{Legal Accountability}}}.
\newblock Oxford University Press, 2020.
\newblock ISBN 978-0-19-883606-3.

\bibitem[Kolieb and Poblet(2018)]{koliebRespondingHumanRights2018}
Jonathan Kolieb and Marta Poblet.
\newblock Responding to {Human} {Rights} {Abuses} in the {Digital} {Era}: {New}
  {Tools}, {Old} {Challenges}.
\newblock \emph{Stanford Journal of International Law}, 52\penalty0 (2), 2018.
\newblock URL \url{https://papers.ssrn.com/abstract=3859873}.

\bibitem[Lazer et~al.(2020)Lazer, Pentland, Watts, Aral, Athey, Contractor,
  Freelon, Gonzalez-Bailon, King, Margetts, Nelson, Salganik, Strohmaier,
  Vespignani, and Wagner]{lazerComputationalSocialScience2020}
David M.~J. Lazer, Alex Pentland, Duncan~J. Watts, Sinan Aral, Susan Athey,
  Noshir Contractor, Deen Freelon, Sandra Gonzalez-Bailon, Gary King, Helen
  Margetts, Alondra Nelson, Matthew~J. Salganik, Markus Strohmaier, Alessandro
  Vespignani, and Claudia Wagner.
\newblock Computational social science: {Obstacles} and opportunities.
\newblock \emph{Science}, 369\penalty0 (6507):\penalty0 1060--1062, 2020.
\newblock URL \url{https://doi.org/10.1126/science.aaz8170}.

\bibitem[Lemgruber(2022)]{lemgruberTirosNoFuturo2022}
Julita Lemgruber.
\newblock Tiros no futuro: {Impactos} da guerra às drogas na rede municipal de
  educação do {Rio} de {Janeiro}, 2022.
\newblock URL
  \url{https://cesecseguranca.com.br/textodownload/tiros-no-futuro-impactos-da-guerra-as-drogas-na-rede-municipal-de-educacao-do-rio-de-janeiro/}.

\bibitem[Lozovatsky and Saha(2014)]{lozovatskyImpactFirearmViolence2014}
Michael Lozovatsky and Subrata Saha.
\newblock The {{Impact}} of {{Firearm Violence}} on the {{Healthcare System}}
  of the {{United States}}.
\newblock 5\penalty0 (1), 2014.
\newblock ISSN 2151-805X, 2151-8068.
\newblock \doi{10.1615/EthicsBiologyEngMed.2014012035}.
\newblock URL
  \url{https://www.dl.begellhouse.com/journals/6ed509641f7324e6,59bf2d6a63e1cf43,590e9a6534ab5114.html}.

\bibitem[McClintock(2010)]{mcclintock_standard_2010}
Michael McClintock.
\newblock The {Standard} {Approach} to {Human} {Rights} {Research}.
\newblock In \emph{Human {Rights}: {From} {Practice} to {Policy}}. Gerald R.
  Ford School of Public Policy University of Michigan, 2010.

\bibitem[Miller et~al.(2013)Miller, Shrestha, Derby, Olive, Umapathy, Li, and
  Zhao]{millerDiggingHumanRights2013}
Ben Miller, Ayush Shrestha, Jason Derby, Jennifer Olive, Karthikeyan Umapathy,
  Fuxin Li, and Yanjun Zhao.
\newblock Digging into human rights violations: {Data} modelling and collective
  memory.
\newblock In \emph{2013 {IEEE} {International} {Conference} on {Big} {Data}},
  pages 37--45, 2013.
\newblock URL \url{https://doi.org/10.1109/BigData.2013.6691668}.

\bibitem[Mooney et~al.(2022)Mooney, Pundik, Raymond, and
  Simon]{mooneySocialMediaEvidence2021}
Olivia Mooney, Kate Pundik, Nathaniel Raymond, and David Simon.
\newblock Social {{Media Evidence}} of {{Alleged Gross Human Rights Abuses}}:
  {{Improving Preservation}} and {{Access Through Policy Reform}}, 2022.
\newblock URL
  \url{https://gsp.yale.edu/social-media-evidence-alleged-gross-human-rights-abuses-improving-preservation-and-access-through}.

\bibitem[Murdie and Watson(2021)]{murdie_quantitative_2021}
Amanda~M. Murdie and K.~Anne Watson.
\newblock Quantitative {Human} {Rights}.
\newblock \emph{Oxford Research Encyclopedia of International Studies}, 2021.
\newblock URL \url{https://doi.org/10.1093/acrefore/9780190846626.013.603}.

\bibitem[Myers(2020)]{myersHowConductDiscovery2020}
Paul Myers.
\newblock \emph{How to {{Conduct Discovery Using Open Source Methods}}}.
\newblock Oxford University Press, 2020.
\newblock ISBN 978-0-19-883606-3.

\bibitem[Namisango et~al.(2019)Namisango, Kang, and
  Rehman]{namisangoWhatWeKnow2019}
Fatuma Namisango, Kyeong Kang, and Junaid Rehman.
\newblock What {{Do We Know}} about {{Social Media}} in {{Nonprofits}}? {{A
  Review}}.
\newblock 2019.

\bibitem[Nations(2013)]{unitednationsHumanRightsIndicators2013}
United Nations.
\newblock Human {Rights} {Indicators}: {A} {Guide} to {Measurement} and
  {Implementation}, 2013.
\newblock URL \url{https://doi.org/10.18356/58576336-en}.

\bibitem[Nemkova et~al.(2023)Nemkova, Ubani, Polat, Kim, and
  Nielsen]{nemkovaDetectingHumanRights2023}
Poli Nemkova, Solomon Ubani, Suleyman~Olcay Polat, Nayeon Kim, and Rodney~D.
  Nielsen.
\newblock Detecting {Human} {Rights} {Violations} on {Social} {Media} during
  {Russia}-{Ukraine} {War}, 2023.
\newblock URL \url{https://doi.org/10.48550/arXiv.2306.05370}.

\bibitem[Nielsen et~al.(2024)Nielsen, Landwehr, Nicolaï, Patil, and
  Raju]{nielsenSocialMediaCrowdsourcing2024}
Anne~B. Nielsen, Dario Landwehr, Juliette Nicolaï, Tejal Patil, and Emmanuel
  Raju.
\newblock Social media and crowdsourcing in disaster risk management:
  {{Trends}}, gaps, and insights from the current state of research.
\newblock 15\penalty0 (2):\penalty0 104--127, 2024.
\newblock ISSN 1944-4079, 1944-4079.
\newblock \doi{10.1002/rhc3.12297}.
\newblock URL \url{https://onlinelibrary.wiley.com/doi/10.1002/rhc3.12297}.

\bibitem[of~Disease 2016 Injury~Collaborators(2018)]{globalBurden2018}
The Global~Burden of~Disease 2016 Injury~Collaborators.
\newblock Global {Mortality} {From} {Firearms}, 1990-2016.
\newblock \emph{JAMA}, 320\penalty0 (8):\penalty0 792, 2018.
\newblock ISSN 0098-7484.
\newblock URL \url{https://doi.org/10.1001/jama.2018.10060}.

\bibitem[Ou et~al.(2022)Ou, Ren, Duan, Tang, Zhu, Feng, Zhang, Liang, Su,
  Zhang, Cui, Chen, Zhou, Mao, and Wang]{ouGlobalBurdenTrends2022}
Zejin Ou, Yixian Ren, Danping Duan, Shihao Tang, Shaofang Zhu, Kexin Feng,
  Jinwei Zhang, Jiabin Liang, Yiwei Su, Yuxia Zhang, Jiaxin Cui, Yuquan Chen,
  Xueqiong Zhou, Chen Mao, and Zhi Wang.
\newblock Global burden and trends of firearm violence in 204
  countries/territories from 1990 to 2019.
\newblock \emph{Frontiers in Public Health}, 10, 2022.
\newblock ISSN 2296-2565.
\newblock \doi{10.3389/fpubh.2022.966507}.
\newblock URL
  \url{https://www.frontiersin.org/articles/10.3389/fpubh.2022.966507}.

\bibitem[Pavlick and Callison-Burch(2016)]{pavlickGunViolenceDatabase2016}
Ellie Pavlick and Chris Callison-Burch.
\newblock The gun violence database.
\newblock In \emph{Presented at the {Data} {For} {Good} {Exchange} 2016}, 2016.
\newblock URL \url{https://doi.org/10.48550/arXiv.1610.01670}.

\bibitem[Pilankar et~al.(2022)Pilankar, Haque, Hasanuzzaman, Stynes, and
  Pathak]{pilankarDetectingViolationHuman2022}
Yash Pilankar, Rejwanul Haque, Mohammed Hasanuzzaman, Paul Stynes, and Pramod
  Pathak.
\newblock Detecting {Violation} of {Human} {Rights} via {Social} {Media}.
\newblock In \emph{Proceedings of the {First} {Computing} {Social}
  {Responsibility} {Workshop} within the 13th {Language} {Resources} and
  {Evaluation} {Conference}}, pages 40--45. European Language Resources
  Association, 2022.
\newblock URL \url{https://aclanthology.org/2022.csrnlp-1.6}.

\bibitem[Piskorski et~al.(2020)Piskorski, Haneczok, and
  Jacquet]{piskorski_new_2020}
Jakub Piskorski, Jacek Haneczok, and Guillaume Jacquet.
\newblock New {Benchmark} {Corpus} and {Models} for {Fine}-grained {Event}
  {Classification}: {To} {BERT} or not to {BERT}?
\newblock In \emph{Proceedings of the 28th {International} {Conference} on
  {Computational} {Linguistics}}, pages 6663--6678. International Committee on
  Computational Linguistics, 2020.
\newblock URL \url{https://doi.org/10.18653/v1/2020.coling-main.584}.

\bibitem[Price and
  Ball(2015{\natexlab{a}})]{priceLimitsObservationUnderstanding2015}
Megan Price and Patrick Ball.
\newblock The {Limits} of {Observation} for {Understanding} {Mass} {Violence}.
\newblock \emph{Canadian Journal of Law and Society / La Revue Canadienne Droit
  et Société}, 30\penalty0 (2):\penalty0 237--257, 2015{\natexlab{a}}.
\newblock ISSN 0829-3201, 1911-0227.
\newblock URL \url{https://doi.org/10.1017/cls.2015.24}.

\bibitem[Price and Ball(2015{\natexlab{b}})]{price_selection_2015}
Megan Price and Patrick Ball.
\newblock Selection bias and the statistical patterns of mortality in conflict.
\newblock \emph{Statistical Journal of the IAOS}, 31\penalty0 (2):\penalty0
  263--272, 2015{\natexlab{b}}.

\bibitem[Raleigh et~al.(2010)Raleigh, Linke, Hegre, and
  Karlsen]{raleighIntroducingACLEDArmed2010}
Clionadh Raleigh, rew Linke, Håvard Hegre, and Joakim Karlsen.
\newblock Introducing {ACLED}: {An} armed conflict location and event dataset.
\newblock \emph{Journal of peace research}, 47\penalty0 (5):\penalty0 651--660,
  2010.
\newblock URL \url{https://doi.org/10.1177/0022343310378914}.

\bibitem[Ran et~al.(2023)Ran, Lu, Tetreault, Cahill, and
  Jaimes]{ranNewTaskDataset2023}
Shihao Ran, Di~Lu, Joel Tetreault, Aoife Cahill, and Alejandro Jaimes.
\newblock A {New} {Task} and {Dataset} on {Detecting} {Attacks} on {Human}
  {Rights} {Defenders}, 2023.
\newblock URL \url{https://doi.org/10.48550/arXiv.2306.17695}.

\bibitem[Rossman and Rallis(2017)]{rossmanIntroductionQualitativeResearch2017}
Gretchen~B. Rossman and Sharon~F. Rallis.
\newblock \emph{An {Introduction} to {Qualitative} {Research}: {Learning} in
  the {Field}}.
\newblock SAGE Publications, 2017.
\newblock ISBN 978-1-07-180269-4.
\newblock URL \url{https://doi.org/10.4135/9781071802694}.

\bibitem[Shabani and Sokhn(2018)]{shabaniHybridMachineCrowdApproach2018}
Shaban Shabani and Maria Sokhn.
\newblock Hybrid {{Machine-Crowd Approach}} for {{Fake News Detection}}.
\newblock In \emph{2018 {{IEEE}} 4th {{International Conference}} on
  {{Collaboration}} and {{Internet Computing}} ({{CIC}})}, pages 299--306,
  2018.
\newblock \doi{10.1109/CIC.2018.00048}.
\newblock URL \url{https://ieeexplore.ieee.org/abstract/document/8537846}.

\bibitem[Silva et~al.(2021)Silva, Ribeiro, Frossard, Souza, Schenker, and
  Minayo]{silvaNoMeioFogo2021}
Mayalu~Matos Silva, Fernanda Mendes~Lages Ribeiro, Vera~Cecília Frossard,
  Rosane Marques~de Souza, Miriam Schenker, and Maria Cecília de~Souza Minayo.
\newblock “{No} meio do fogo cruzado”: reflexões sobre os impactos da
  violência armada na {Atenção} {Primária} em {Saúde} no município do
  {Rio} de {Janeiro}.
\newblock \emph{Ciência \& Saúde Coletiva}, 26:\penalty0 2109--2118, 2021.
\newblock ISSN 1413-8123, 1678-4561.
\newblock URL \url{https://doi.org/10.1590/1413-81232021266.00632021}.

\bibitem[Silver et~al.(2023)Silver, Ramos, Stamm, Gladden, Martin, and
  Mulcahey]{silverExaminingHealthcareEconomic2023}
Julia~H. Silver, Tolulope~A. Ramos, Michaela~A. Stamm, Paul~B. Gladden,
  Murphy~P. Martin, and Mary~K. Mulcahey.
\newblock Examining the {{Healthcare}} and {{Economic Burden}} of {{Gun
  Violence}} in a {{Major US Metropolitan City}}.
\newblock 7\penalty0 (8), 2023.
\newblock ISSN 2474-7661.
\newblock \doi{10.5435/JAAOSGlobal-D-22-00158}.
\newblock URL \url{https://www.ncbi.nlm.nih.gov/pmc/articles/PMC10412425/}.

\bibitem[Souza et~al.(2020)Souza, Nogueira, and
  Lotufo]{souzaBERTimbauPretrainedBERT2020}
Fábio Souza, Rodrigo Nogueira, and Roberto Lotufo.
\newblock {BERTimbau}: {Pretrained} {BERT} {Models} for {Brazilian}
  {Portuguese}.
\newblock In Ricardo Cerri and Ronaldo~C. Prati, editors, \emph{Intelligent
  {Systems}}, Lecture {Notes} in {Computer} {Science}, pages 403--417. Springer
  International Publishing, 2020.
\newblock ISBN 978-3-030-61377-8.
\newblock URL \url{https://doi.org/10.1007/978-3-030-61377-8_28}.

\bibitem[Szwarcwald and Castilho(1998)]{szwarcwaldMortalidadePorArmas1998}
Célia~Landman Szwarcwald and Euclides~Ayres Castilho.
\newblock Mortalidade por armas de fogo no estado do {{Rio}} de {{Janeiro}},
  {{Brasil}}: Uma análise espacial.
\newblock 1998.
\newblock URL \url{https://iris.paho.org/handle/10665.2/7802}.

\bibitem[Ta et~al.(2022)Ta, Rahman, Najjar, and
  Gelbukh]{taGANBERTAdversarialLearning2022}
Hoang~Thang Ta, Abu Bakar~Siddiqur Rahman, Lotfollah Najjar, and Alexander
  Gelbukh.
\newblock {GAN}-{BERT}: {Adversarial} {Learning} for {Detection} of
  {Aggressive} and {Violent} {Incidents} from {Social} {Media}.
\newblock In \emph{Proceedings of the {Iberian} {Languages} {Evaluation}
  {Forum} ({IberLEF} 2022), {CEUR} {Workshop} {Proceedings}}, 2022.
\newblock URL \url{https://ceur-ws.org/Vol-3202/davincis-paper7.pdf}.

\bibitem[Vaswani et~al.(2017)Vaswani, Shazeer, Parmar, Uszkoreit, Jones, Gomez,
  Kaiser, and Polosukhin]{vaswaniAttentionAllYou2017}
Ashish Vaswani, Noam Shazeer, Niki Parmar, Jakob Uszkoreit, Llion Jones,
  Aidan~N Gomez, Łukasz Kaiser, and Illia Polosukhin.
\newblock Attention is all you need.
\newblock In I.~Guyon, U.~Von Luxburg, S.~Bengio, H.~Wallach, R.~Fergus,
  S.~Vishwanathan, and R.~Garnett, editors, \emph{Advances in neural
  information processing systems}, volume~30, 2017.
\newblock URL
  \url{https://proceedings.neurips.cc/paper_files/paper/2017/file/3f5ee243547dee91fbd053c1c4a845aa-Paper.pdf}.

\bibitem[Wagner~Filho et~al.(2018)Wagner~Filho, Wilkens, Idiart, and
  Villavicencio]{wagnerfilhoBrWaCCorpusNew2018}
Jorge~A. Wagner~Filho, Rodrigo Wilkens, Marco Idiart, and Aline Villavicencio.
\newblock The {brWaC} corpus: a new open resource for {Brazilian} {Portuguese}.
\newblock In \emph{Proceedings of the eleventh international conference on
  language resources and evaluation ({LREC} 2018)}. European Language Resources
  Association, 2018.
\newblock URL \url{https://aclanthology.org/L18-1686}.

\end{thebibliography}
\newpage
\appendix
\section{Model evaluation}
\label{app:model_evaluation}

\begin{figure}[H]
    \centering
    \includegraphics[width=0.7\linewidth]{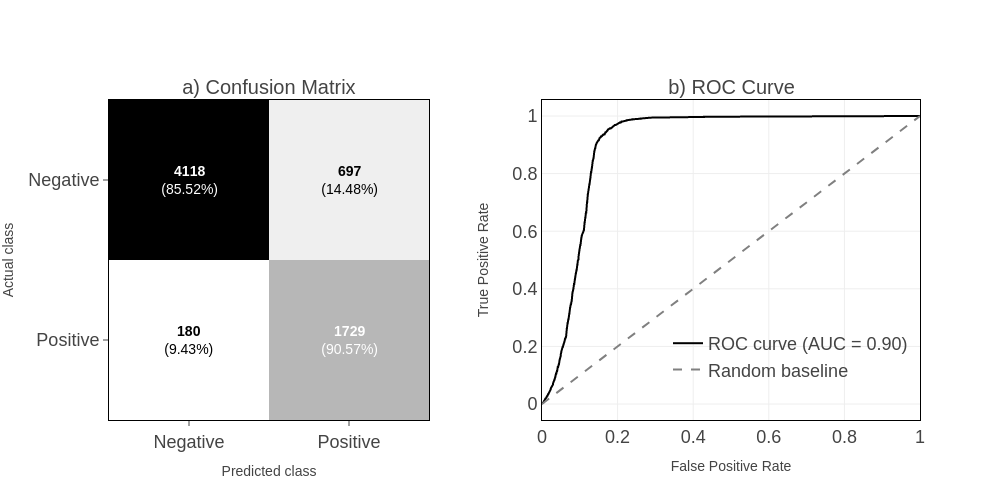}
    \caption{Confusion matrix and ROC curve for BERTimbau using $H_{interactions}$.}
    \label{fig:eval-interactions}
\end{figure}

\begin{figure}[H]
    \centering
    \includegraphics[width=0.7\linewidth]{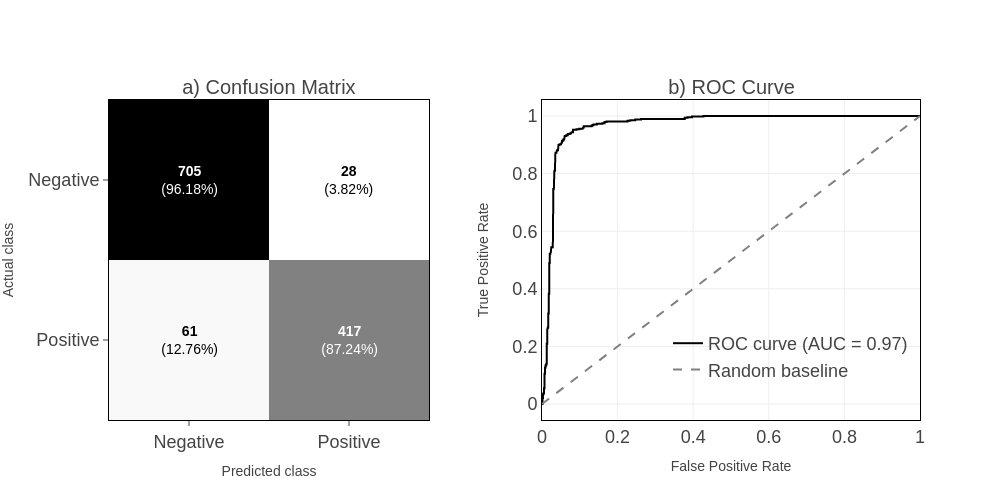}
    \caption{Confusion matrix and ROC curve for BERTimbau using $H_{reports}$.}
    \label{fig:eval-hlabel}
\end{figure}

\section{Interview and survey}
\label{app:results}
\subsection{Interview questions}
One month after starting the intervention, we interviewed four participants from Fogo Cruzado's team in Rio de Janeiro in July 2023 using the following questions. The questions sought to explore the participants' overall experience with the intervention and evaluate the efficacy of the prototype developed. 

\begin{itemize}
\item How do you describe your overall experience using the prototype?
\item Have you used to identify non-geotagged messages?
\item How often did you check the posts classified as negative cases?
\item How have you combined the prototype with your traditional workflow to monitor tweets in Tweetdeck?
\item What were the main advantages of using the prototype?
\item What were the main drawbacks?
\item Can you identify any pattern in cases where the model misclassifies a report?
\item How important are messages that are not potential reports of gun violence but might contain relevant information for your monitoring work?
\end{itemize}

\clearpage

\subsection{Survey results}
\label{app:questionnaire}
We sent the first online survey (Survey \#1) to Fogo Cruzado's staff in Rio de Janeiro on 26 May 26 2023, before the intervention, which started on 29 May 2023. The form to evaluate the intervention (Survey \#2) was shared on 4 September 2023. Four out of the five participants in the core team of analysts in Rio de Janeiro answered the survey. 

\begin{figure}[H]
    \centering
    \includegraphics[width=0.8\linewidth]{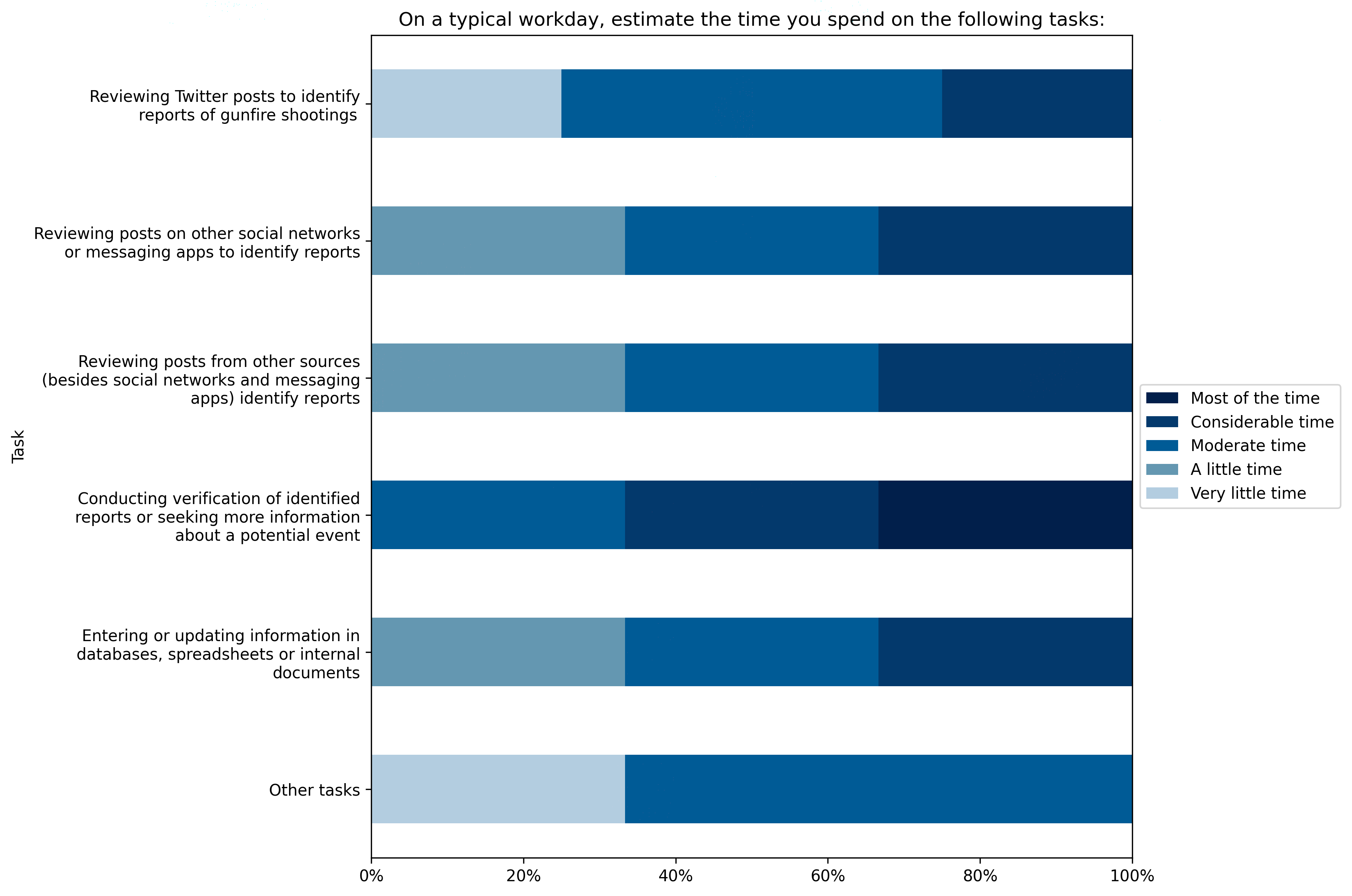}
    \caption{On a typical workday, estimate the time you spend on the following tasks: (question 1).}
    \label{fig:presurvey1}
\end{figure}

\begin{figure}[H]
    \centering
    \includegraphics[width=0.8\linewidth]{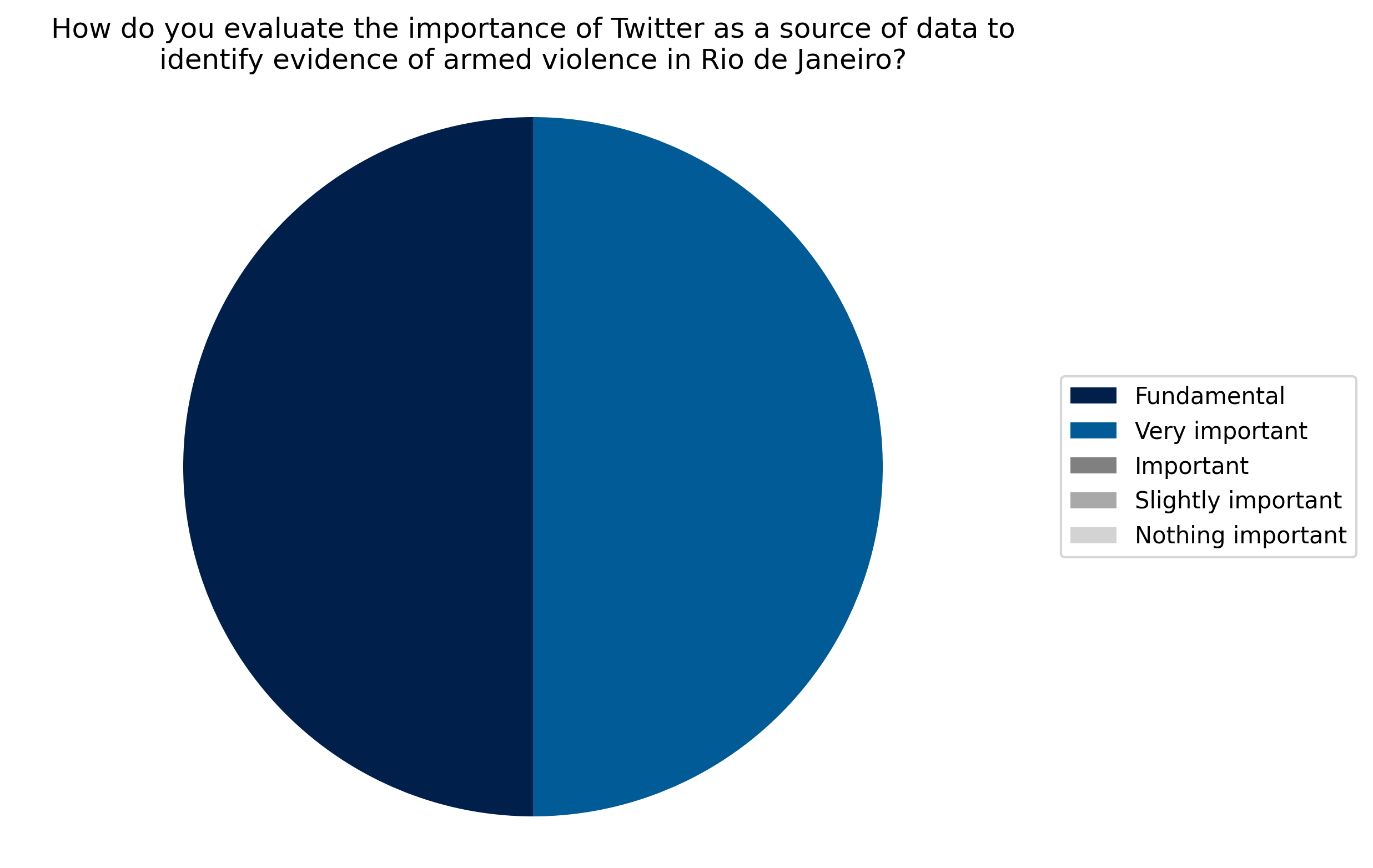}
    \caption{Survey prior to the intervention (question 2).}
    \label{fig:presurvey2}
\end{figure}

\begin{figure}[H]
    \centering
    \includegraphics[width=0.8\linewidth]{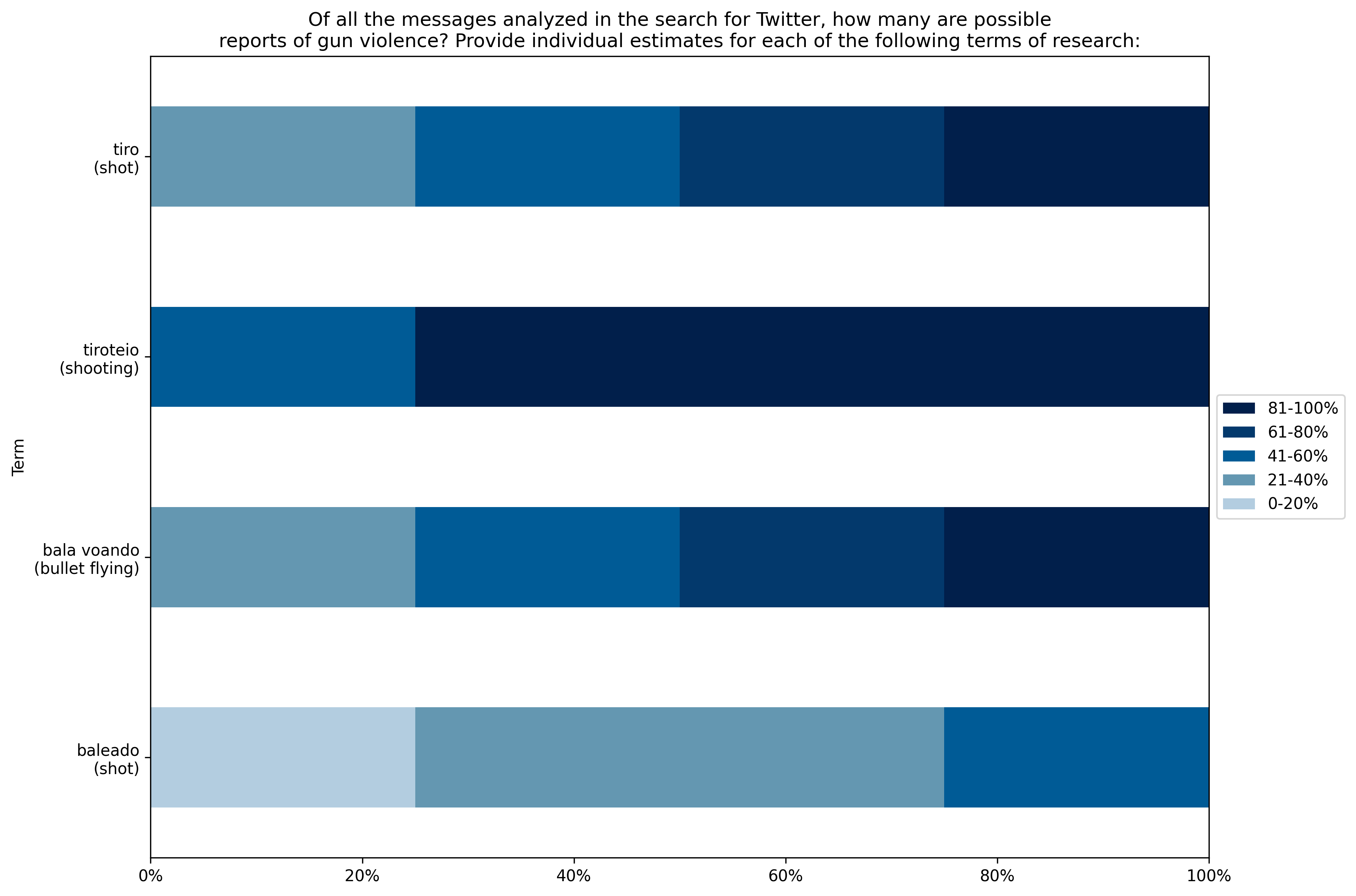}
    \caption{Survey prior to the intervention (question 3).}
    \label{fig:presurvey3}
\end{figure}

\begin{figure}[H]
    \centering
    \includegraphics[width=0.8\linewidth]{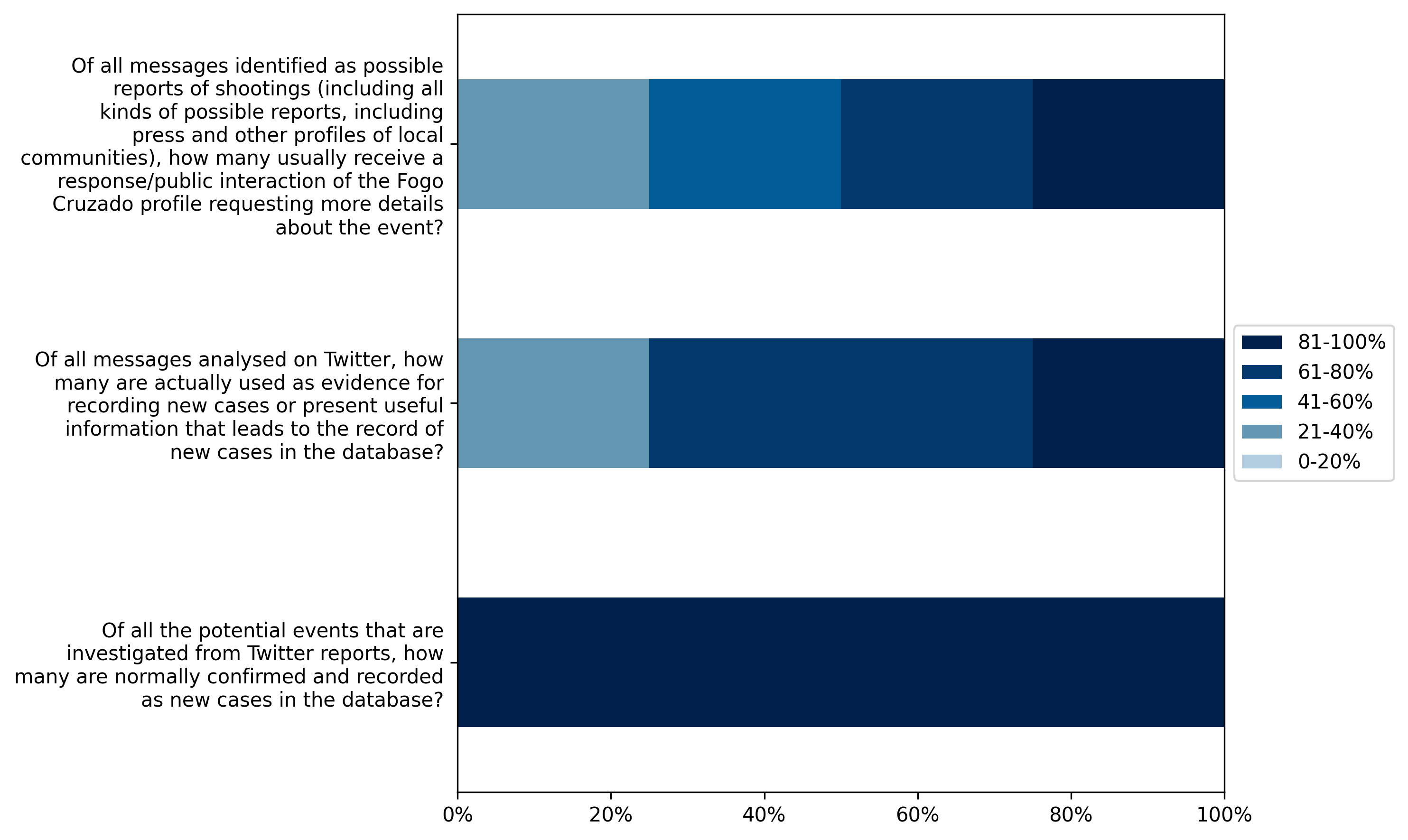}
    \caption{Survey prior to the intervention (questions 4 to 6).}
    \label{fig:presurvey4to6}
\end{figure}

\begin{figure}[H]
    \centering
    \includegraphics[width=0.8\linewidth]{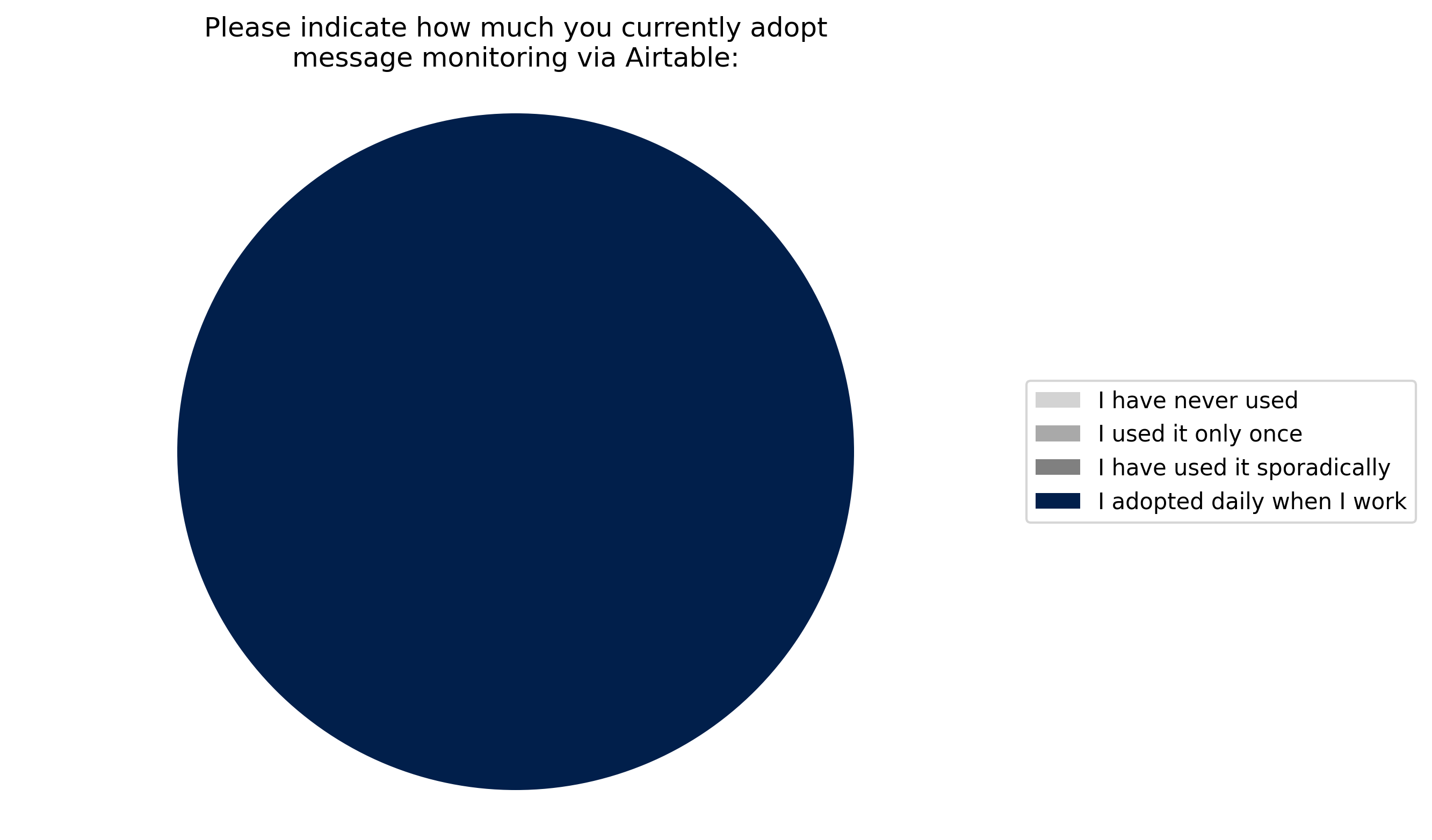}
    \caption{Survey after the intervention (question 1).}
    \label{fig:postsurvey1}
\end{figure}

\begin{figure}[H]
    \centering
    \includegraphics[width=0.8\linewidth]{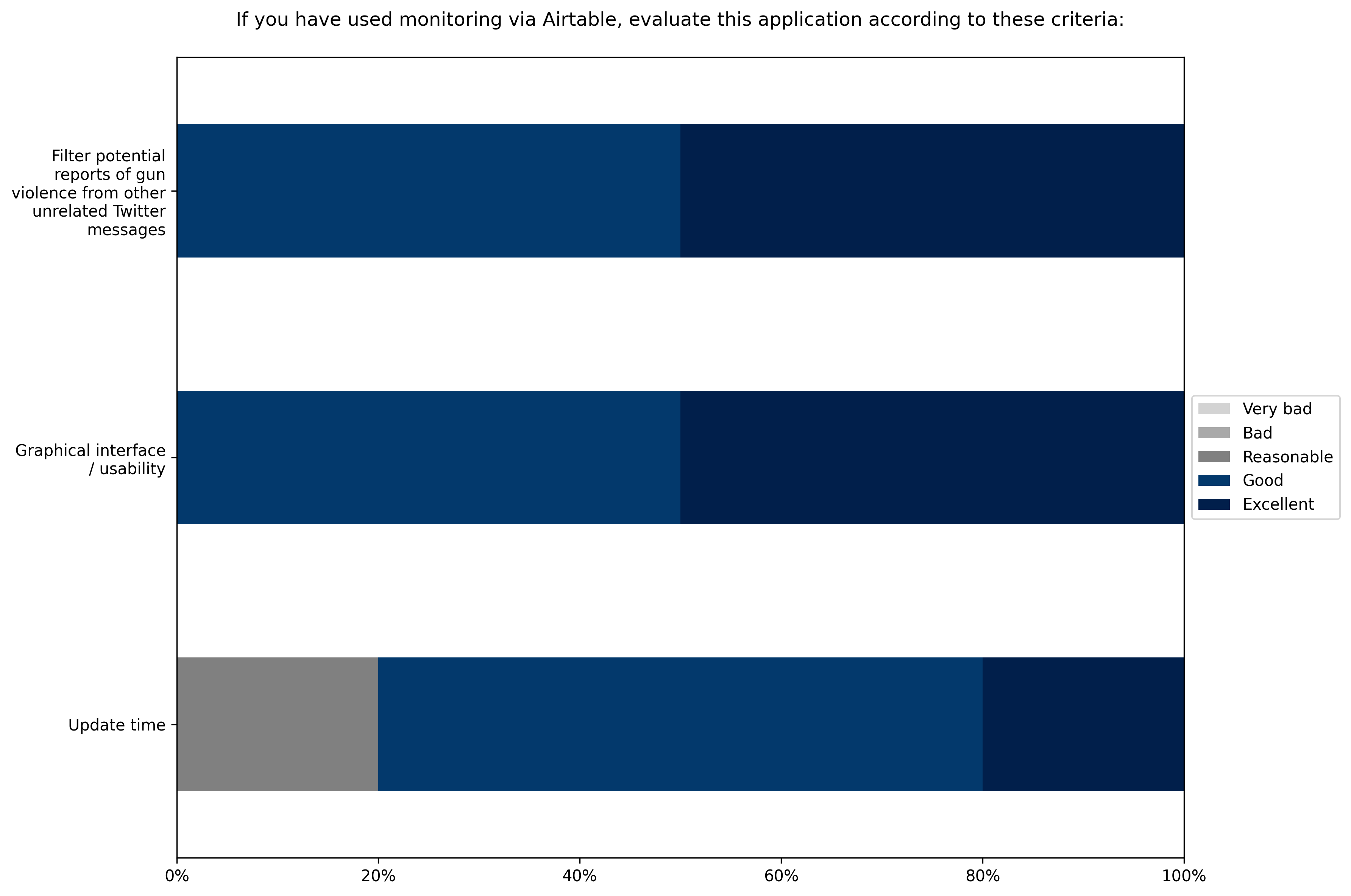}
    \caption{Survey after the intervention (question 2).}
    \label{fig:postsurvey2}
\end{figure}

\begin{figure}[H]
    \centering
    \includegraphics[width=0.8\linewidth]{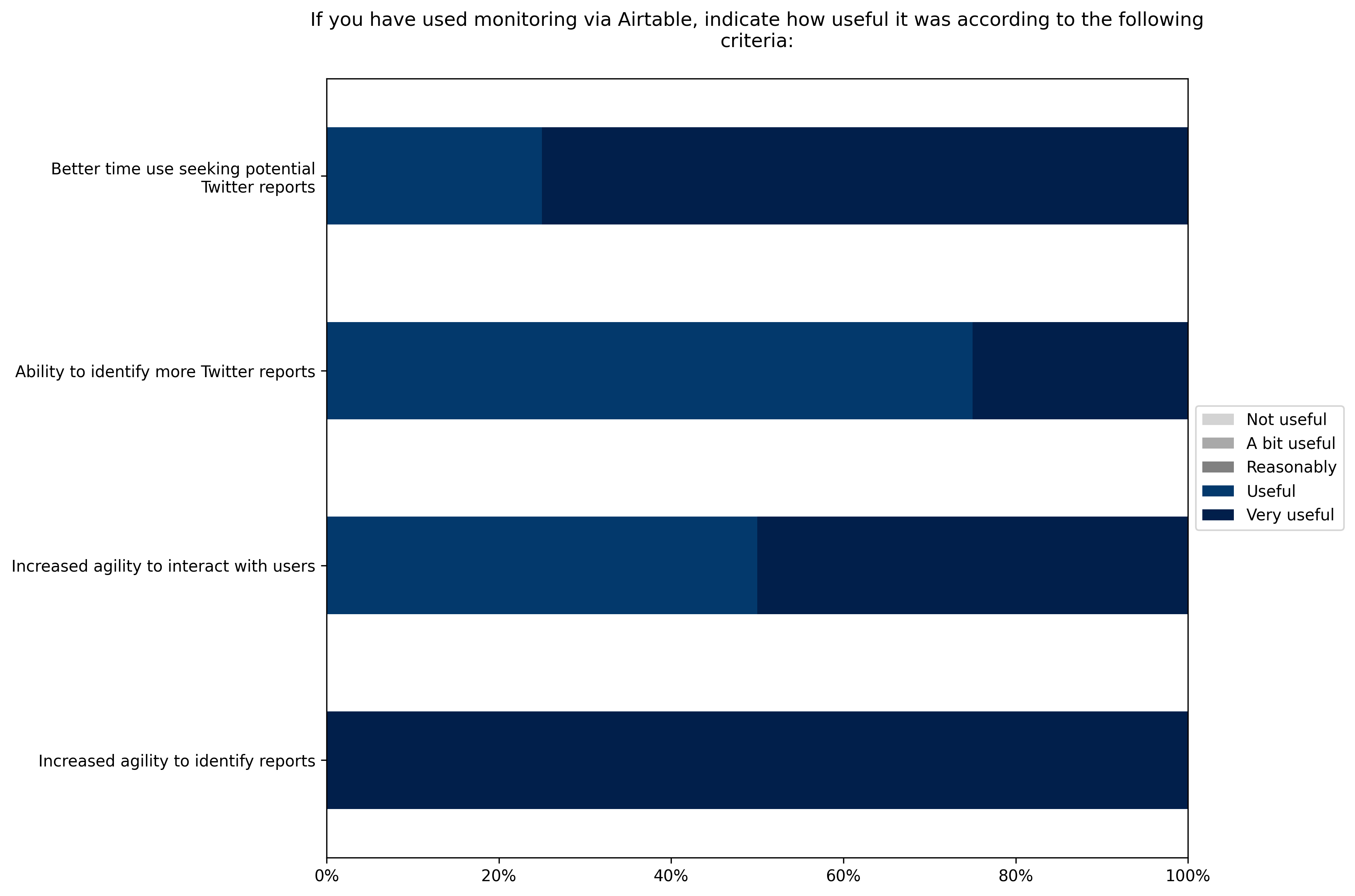}
    \caption{Survey after the intervention (question 3).}
    \label{fig:postsurvey3}
\end{figure}

\begin{figure}[H]
    \centering
    \includegraphics[width=0.8\linewidth]{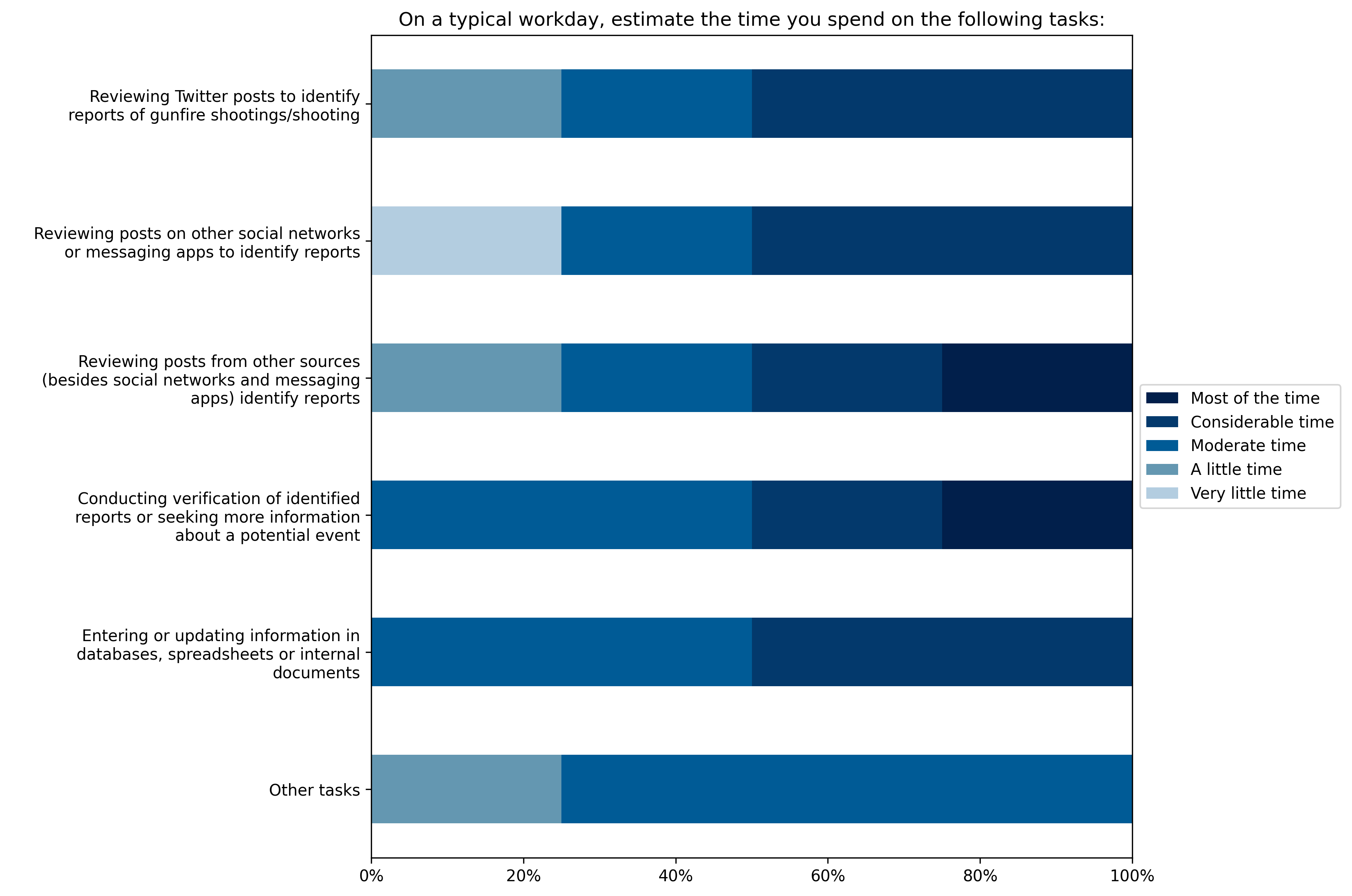}
    \caption{Survey after the intervention (question 4).}
    \label{fig:postsurvey4}
\end{figure}

\begin{figure}[H]
    \centering
    \includegraphics[width=0.8\linewidth]{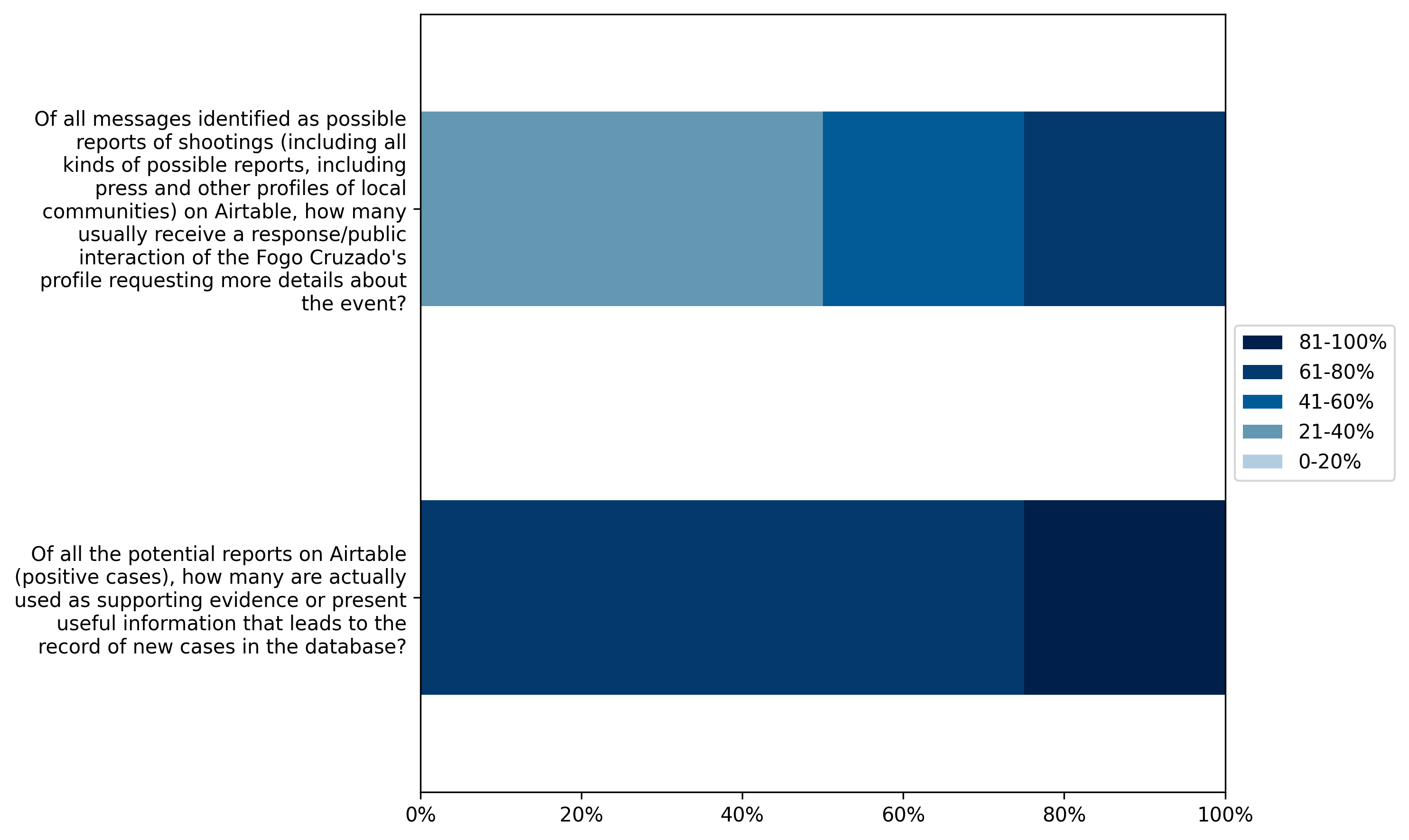}
    \caption{Survey after the intervention (questions 5 and 6).}
    \label{fig:postsurvey5to6}
\end{figure}

\section{Difference-in-difference model}

\label{app:diffindiff}
This Appendix seeks to answer the following question: Did the intervention allow analysts to interact with more messages? The intervention involved changing Fogo Cruzado's workflow for searching and interacting with citizen-generated reports of gun violence on Twitter. First, we developed a Natural Language Processing model to detect gun violence reports and implemented a prototype for real-time monitoring of Twitter messages. Fogo Cruzado's analysts in Rio de Janeiro, Brazil, used this prototype for five weeks. Then, we analyzed difference-in-difference models with control variables and Fogo Cruzado's team in Bahia as a control group. We found evidence suggesting that adopting the model increases analysts' capacity to interact with reports.

We defined the pre-intervention start date as March 1 2023 because it was when the last change in Fogo Cruzado's analyst staff happened; since then, the team composition has remained unchanged. The intervention period under analysis spans from May 29 to July 2 2023. We collected daily data from Fogo Cruzado's public API\footnote{\href{https://api.fogocruzado.org.br/ocurrences}{https://api.fogocruzado.org.br/ocurrences}} and Twitter profiles\footnote{@FogoCruzadoBA, @FogoCruzadoRJ, @FogoCruzadoPE}. The former was used to get records from gun violence events, and the latter measured the number of interactions with gun violence reports.

\begin{figure}[H]
    \centering
    \includegraphics[width=0.8\linewidth]{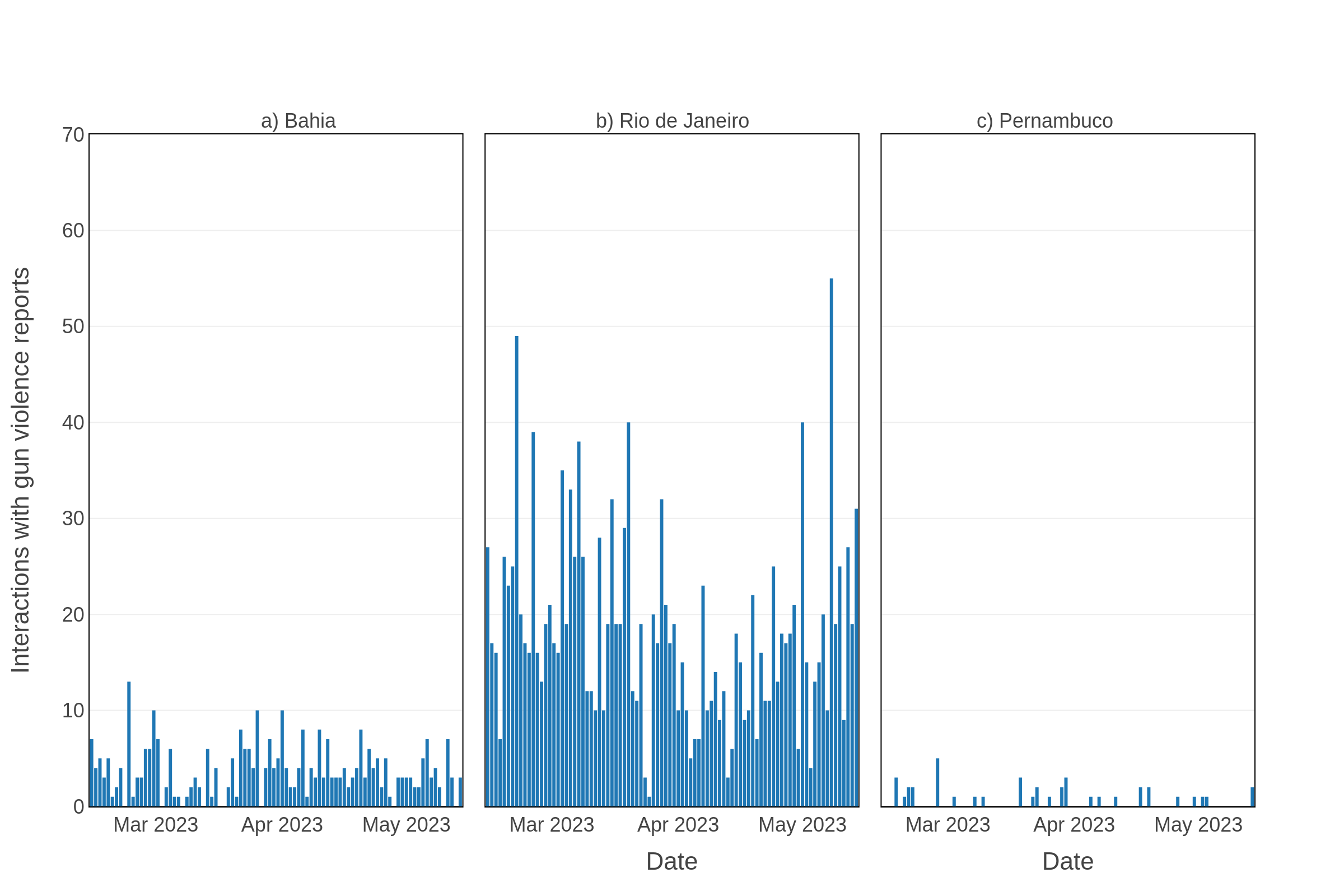}
    \caption{Fogo Cruzado's interactions with users reporting gun violence per day on Twitter in Bahia, Rio de Janeiro, and Pernambuco.}
    \label{fig:interactions}
\end{figure}

Figure~\ref{fig:interactions} presents the total of Fogo Cruzado's interactions with users in all metropolitan regions currently monitored: Rio de Janeiro, Bahia and Pernambuco. Fogo Cruzado uses different Twitter profiles in each region. Although all teams employ the same methodology in all states, the use of Twitter to crowdsource gun violence reports varies across the three states. Several factors can potentially influence the total number of interactions on Twitter across states, such as the number of gun violence events in a single day, their lethality, or demographic differences in the Twitter user base in Brazil. 

Figure~\ref{fig:interactions} shows that while Rio de Janeiro presents an intense activity, Bahia has a much smaller but consistent number of daily replies to gun violence reports. In contrast, Pernambuco has only sporadic activity, with long periods without interactions with users reporting gun violence on Twitter. This characteristic justifies the choice of Bahia as a control group. 

Then, we estimated the impact of the intervention using a difference-in-difference approach. We defined Fogo Cruzado's team in Rio de Janeiro as a treatment group and Bahia as a control group. This research design is possible because both groups use the same methodology to track cases, but Bahia did not have access to the NLP-powered prototype to browse tweets. 

The dependent variable is the count of interactions with gun violence reports on Twitter per day and region/user. The users are \nolinkurl{@FogoCruzadoRJ} and \nolinkurl{@FogoCruzadoBA}, i.e., the Twitter profiles maintained by Fogo Cruzado to monitor and interact with reports from the metropolitan region of two Brazilian states, namely Rio de Janeiro and Bahia. The distribution of the dependent variable is highly right-skewed, with an average of 12 replies and a variance of 116.

The difference in means indicates that the intervention allowed analysts from Rio de Janeiro to interact with nine more reports on average. The mean number of replies before the intervention in Bahia was found to be 6, while in Rio de Janeiro, it was 17. After implementing the intervention, the mean number of replies decreased to 4 (-2) in Bahia, while in Rio de Janeiro, it increased to 24 (+7). The difference in the means is 9; thus, the intervention is associated with an increase of about nine interactions with reports of gun violence per day. We obtained the same estimation using the regression models and controlling for confounding variables. 

\begin{figure}[H]
    \centering
    \includegraphics[width=0.8\linewidth]{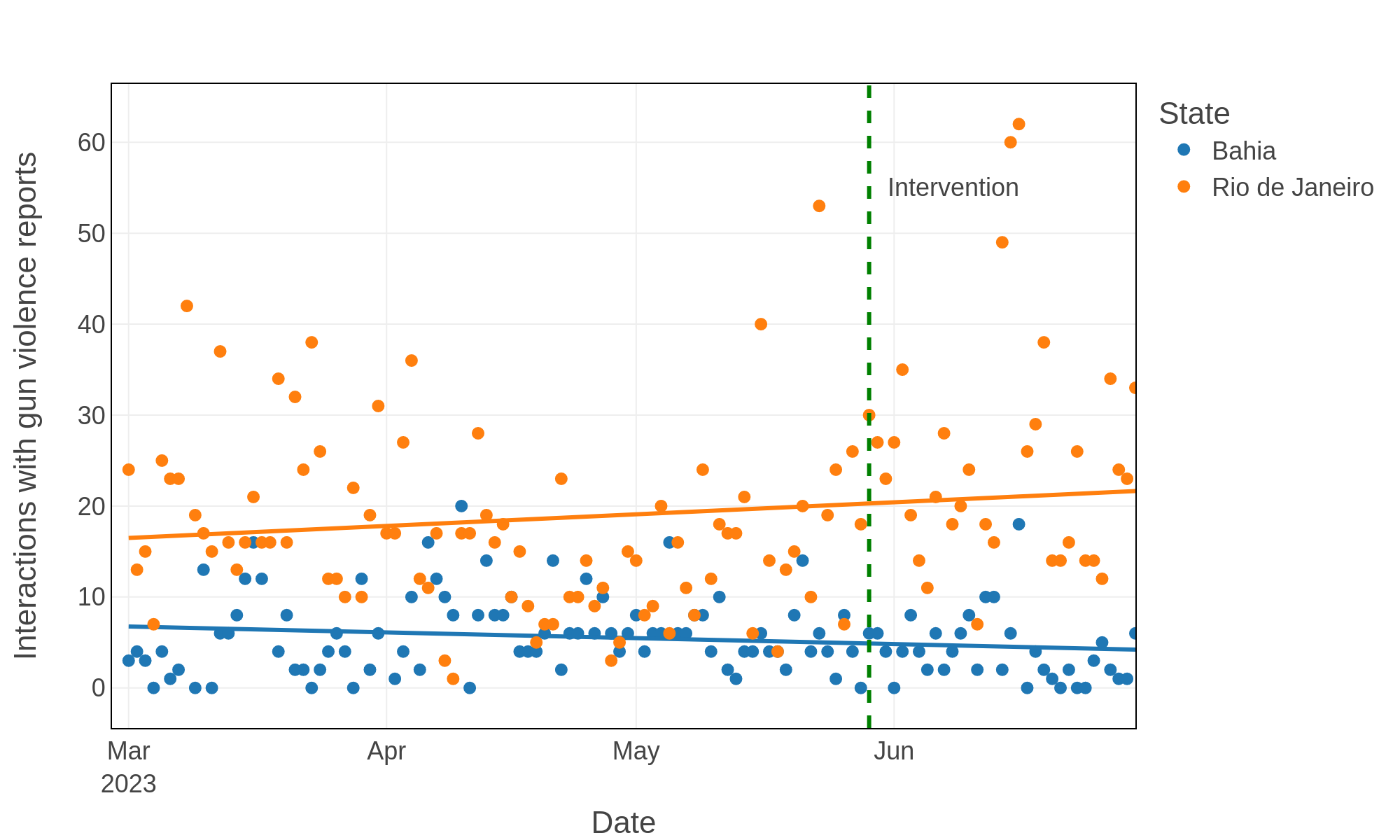}
    \caption{Interactions per day of Fogo Cruzado's team in Bahia and Rio de Janeiro before and after the intervention.}
    \label{fig:trendline-after}
\end{figure}

Figure~\ref{fig:trendline-after} shows the number of interactions per day in the treatment (Rio de Janeiro) and control group (Bahia). The chart shows a stable trendline for Bahia and a positive trendline for Rio de Janeiro, reverting the downward trend observed in the pre-intervention period only, which can be observed in Figure~\ref{fig:diffindiff}.

\subsection{Regression models}
We used two regression models in the difference-in-difference analysis: an \textbf{Ordinary Least Square} model and a \textbf{Negative Binomial} model. The first offers a starting point for the analysis and comparison but is not optimal for the task at hand because some of its assumptions may not hold true for skewed count data. Conversely, the Negative Binomial is an alternative for Poisson, which is suitable for cases where the variance is greater than the mean (overdispersed data). 

We defined the following variables to run the difference-in-difference regression. The subscript ($_t$) represents the day. 

\begin{itemize}
    \item \text{$replies_t$}: the dependent variable is the number of interactions with reports of gun violence per day on Twitter.
    \item $\beta_0$: the intercept of the model.
    \item $\beta_1 \cdot intervention_t$: 1 whether the intervention was running in the day $_t$ and 0 otherwise.
    \item $\beta_2 \cdot treatment$: 1 for Rio de Janeiro and 0 for Bahia, indicating the treatment and control group.
    \item $\beta_3 \cdot (intervention_t \cdot treatment_t)$: interaction term between intervention and treatment.
    \item $\beta_4 \cdot number\_cases_t $: number of gun violence events in the day according to Fogo Cruzado's database.
    \item $\beta_5 \cdot number\_victims_t$: number of victims (killed or injured people) of gun violence in the day $_t$  according to Fogo Cruzado's database.
    \item $\beta_6  \cdot avg\_population_t$: population average of the cities affected in the day according to Fogo Cruzado's database.
    \item $\varepsilon_t$: error term.
\end{itemize}

Table~\ref{regression-results} summarises the coefficients for each estimate; values inside parenthesis are the confidence intervals. The formula used for regression was:

\vspace{10mm}
$replies_t = \beta_0 + \beta_1 \cdot intervention_t + \beta_2 \cdot treatment + \beta_3 \cdot (intervention_t \cdot treatment_t) + \beta_4 \cdot number\_cases_t + \beta_5 \cdot number\_victims_t + \beta_6 \cdot avg\_population_t + \varepsilon_t$.
\vspace{10mm}

Both models indicate that the intervention for the treatment group (\textit{intervention\_treatment}) appears to have a statistically significant effect on increasing the number of interactions with gun violence reports on Twitter. The coefficient for this variable (\textit{intervention\_treatment}) in the OLS model is 9.7, an estimate close to the simple difference in means reported before. 

The treatment variable also presents a high coefficient, indicating that being from the treatment group increases the number of interactions per day, even in the absence of the intervention. This finding is coherent with Figure~\ref{fig:interactions}, which shows that Rio de Janeiro has consistently higher interaction than Bahia. There is no statistically significant effect in any of the control variables.

The observed R-squared values fall within a moderate range, specifically from 0.31 to 0.46. These values suggest that the independent variables in the models can predict approximately 31\% to 46\% of the variability in the dependent variable. In the case of the OLS model, nearly 46\% of the fluctuations in \textit{'replies\_t'} can be explained by the model. When we adjust for the number of predictors, the value marginally decreases to 45\%. The Negative Binomial model does not provide a traditional R-squared value. Instead, we use the Pseudo R-squared measure, which is tailored to assess goodness-of-fit in non-linear regression contexts. The Pseudo R-squared value (Cox-Snell) in this model is 0.31.

\begin{table}[!htbp] \centering
\begin{tabular*}{\textwidth}{@{\extracolsep{\fill}} l@{\hspace{20pt}} c c}
\\[-1.8ex]\hline
\hline \\[-1.8ex]
& \multicolumn{2}{c}{Dependent variable: $replies_t$} \
\cr \cline{2-3}
\\[-1.8ex] & \multicolumn{1}{c}{Ordinary Least Square} & \multicolumn{1}{c}{Negative Binomial}  \\
\\[-1.8ex] & (1) & (2) \\
\hline \\[-1.8ex]
 Intercept & 9.430$^{***}$ & 1.986$^{***}$ \\
  & (3.914 , 14.946) & (1.269 , 2.703) \\
 avg\_population & -0.000$^{*}$ & -0.000$^{}$ \\
  & (-0.000 , 0.000) & (-0.000 , 0.000) \\
 intervention & -2.389$^{}$ & -0.454$^{**}$ \\
  & (-5.572 , 0.793) & (-0.892 , -0.017) \\
 intervention\_treatment & 9.693$^{***}$ & 0.832$^{***}$ \\
  & (5.180 , 14.207) & (0.234 , 1.429) \\
 number\_cases & -0.898$^{}$ & -0.039$^{}$ \\
  & (-2.373 , 0.576) & (-0.231 , 0.152) \\
 number\_victims & 0.213$^{}$ & 0.012$^{}$ \\
  & (-0.111 , 0.536) & (-0.031 , 0.054) \\
 treatment & 13.666$^{***}$ & 1.175$^{***}$ \\
  & (9.295 , 18.037) & (0.607 , 1.743) \\
\hline \\[-1.8ex]
 Observations & 245 & 245 \\
 $R^2$ & 0.463 &  \\
 Adjusted $R^2$ & 0.449 &  \\
 Pseudo $R^2$ (Cox-Snell) & & 0.31 \\
 Residual Std. Error & 8.024(df = 238) & 1.000(df = 238)  \\
 F Statistic & 34.203$^{***}$ (df = 6.0; 238.0) & $^{}$ (df = 6; 238) \\
\hline
\hline \\[-1.8ex]
\textit{Note:} & \multicolumn{2}{r}{$^{*}$p$<$0.1; $^{**}$p$<$0.05; $^{***}$p$<$0.01} \\
\end{tabular*}
\caption{\label{regression-results}Difference-in-difference regression results.}
\end{table}
\clearpage
\subsection{Regression diagnostics}
\begin{figure}[H]
    \centering
    \includegraphics[width=0.8\linewidth]{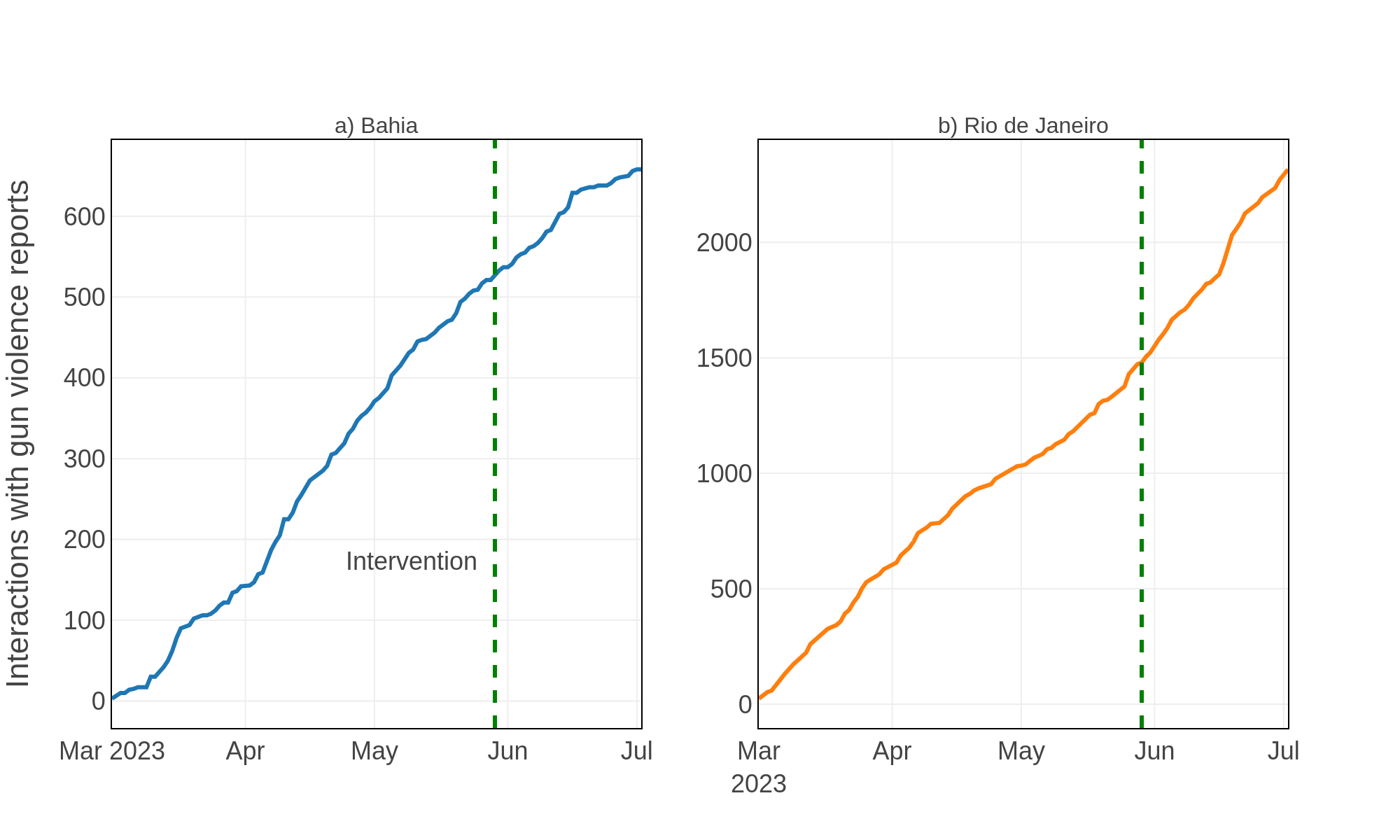}
    \caption{Testing the parallel trend assumption before and after the intervention (cumulative sum of interactions).}
    \label{fig:cumsum-diffindiff}
\end{figure}

\begin{figure}[H]
    \centering
    \includegraphics[width=0.8\linewidth]{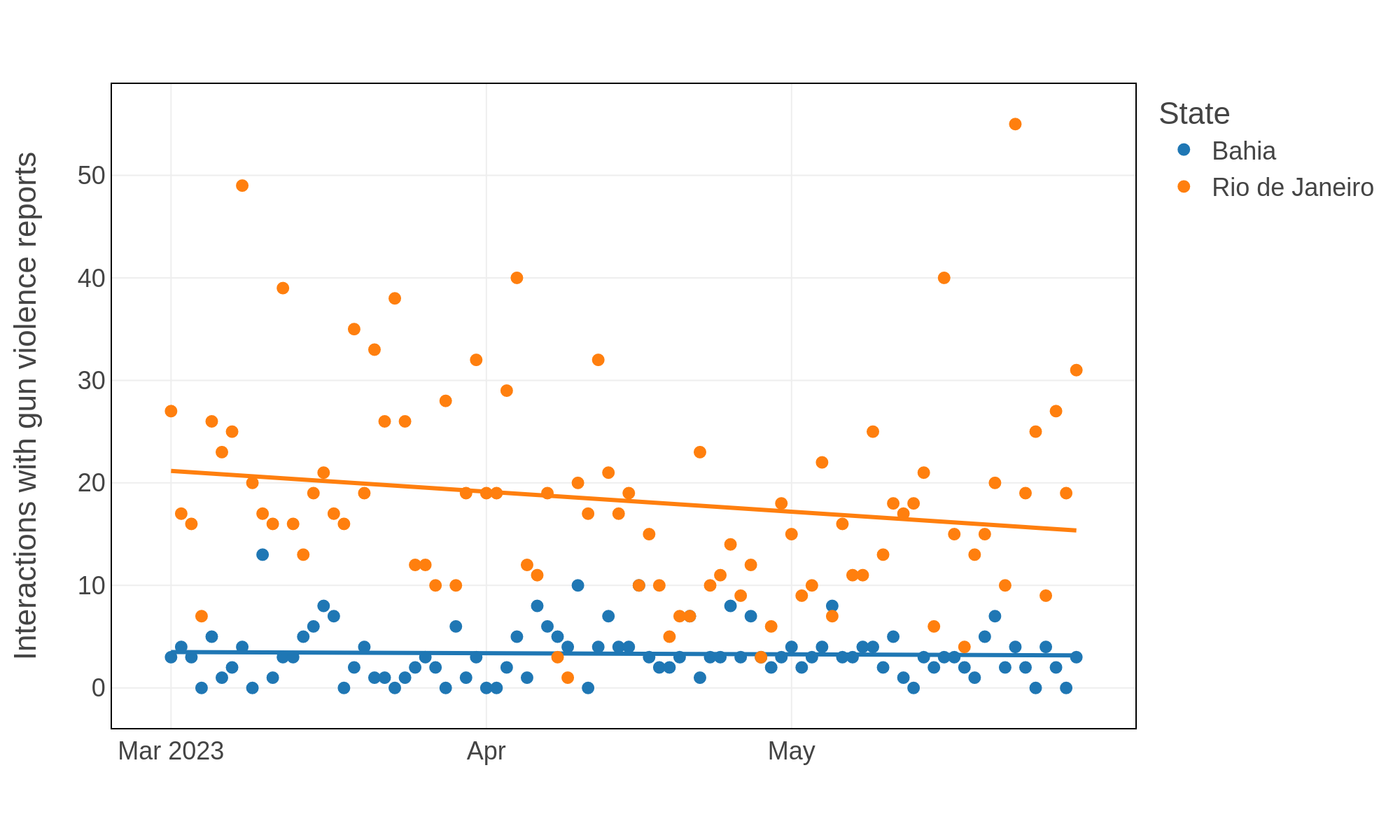}
    \caption{Testing the parallel trend assumption before the intervention (number of interactions per day).}
    \label{fig:diffindiff}
\end{figure}

We used two charts to analyse visually the parallel trend assumption of difference-in-difference models. The parallel trend assumption requires that the changes over time in the dependent variable for both groups are parallel in the absence of the intervention. Figure~\ref{fig:diffindiff} shows the number of daily interactions, i.e., replies to gun violence reports. Figure~\ref{fig:cumsum-diffindiff} shows the growth trend for the cumulative sum of interactions on Twitter. 

There is a rough parallel trend in Figure~\ref{fig:cumsum-diffindiff}. On the other hand, in Figure~\ref{fig:diffindiff}, Rio de Janeiro has a downward trendline, and the trend of daily interactions of Fogo Cruzado's team in Bahia seems stable. 

\begin{figure}[H]
\includegraphics[width=0.8\textwidth]{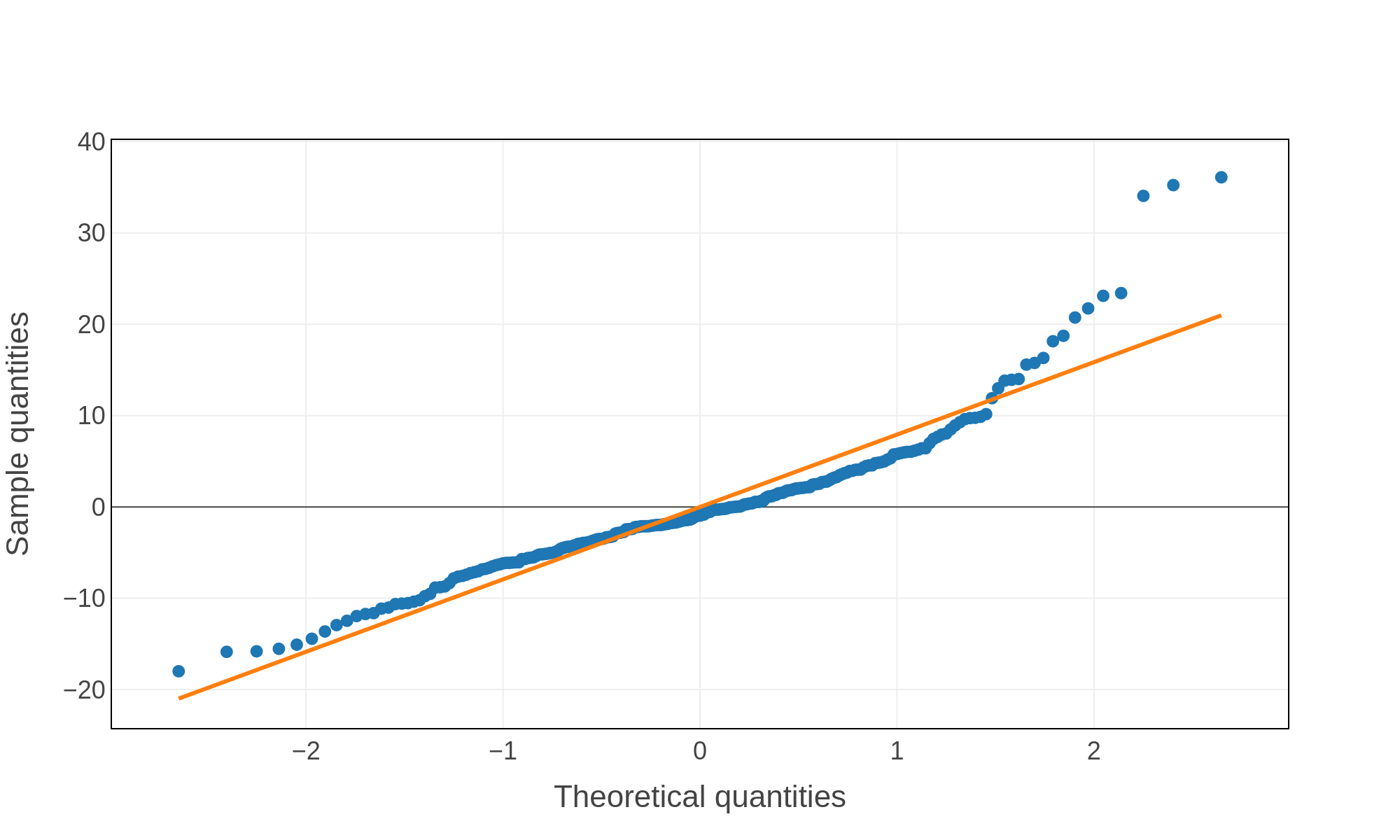}
\caption{QQplot for the regression of the negative binomial model.}
\label{fig:qqplot-neg}
\end{figure}

We assessed the goodness of fit for the models using Quantile-Quantile (QQ) plots, represented in Figure~\ref{fig:qqplot-neg}. The two QQ plots show that points closely align with the theoretical expectations at the centre, which implies that the residuals are approximately normally distributed around the mean, but there are deviations from the theoretical line for extreme quantiles. This pattern suggests that the models have good accuracy for predicting the dependent variable around its mean value but might not be as effective when dealing with the outliers.

\begin{figure}[H]
\centering
\includegraphics[width=0.8\textwidth]{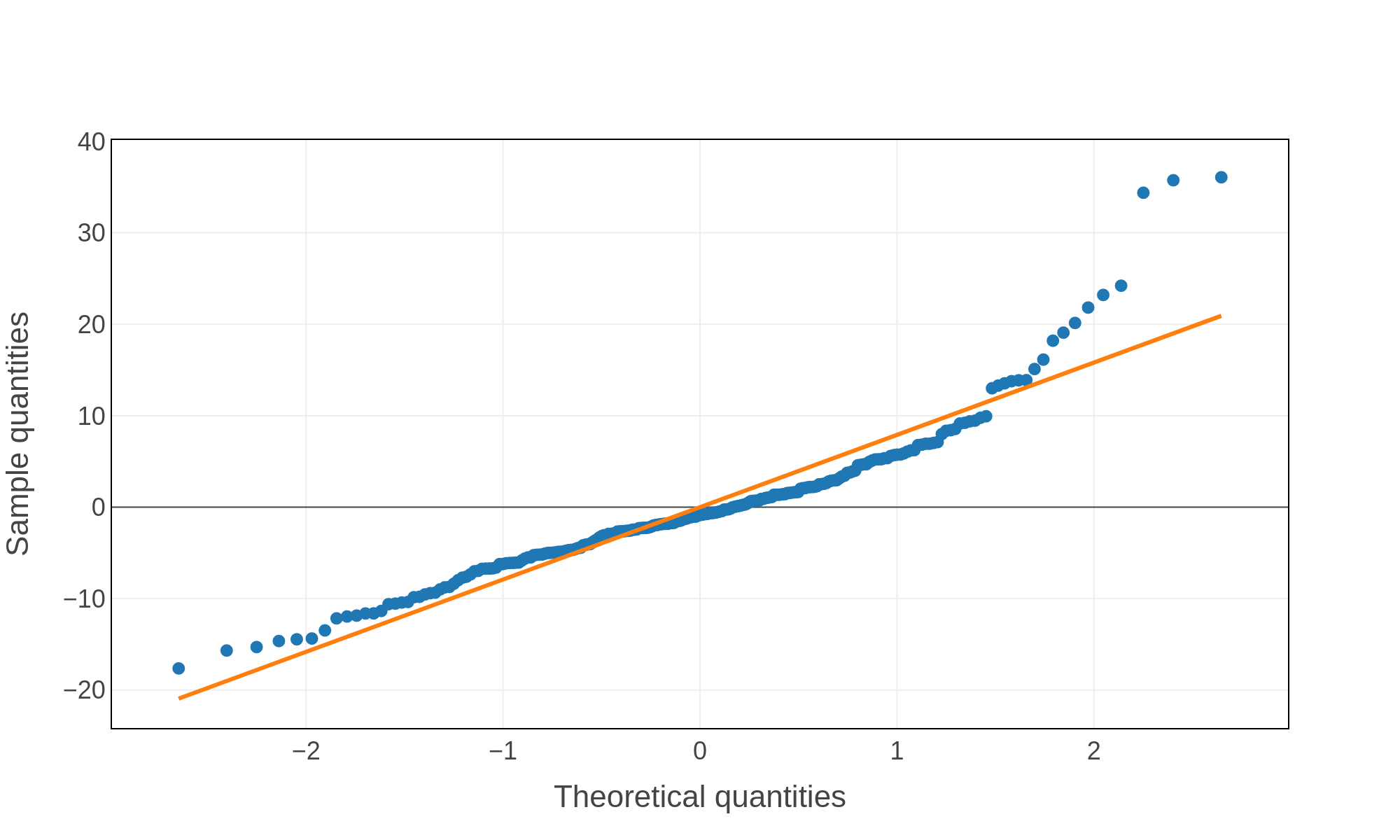}
\caption{QQplot for the Ordinary Least Square model.}
\hfill
\label{fig:qqplot-ols}
\end{figure}    


\end{document}